\documentclass[journal]{IEEEtran}
\usepackage{amsmath,amsfonts}
\usepackage{algorithmic}
\usepackage{algorithm}
\usepackage{array}
\usepackage[caption=false,font=normalsize,labelfont=sf,textfont=sf]{subfig}
\usepackage{textcomp}
\usepackage{stfloats}
\usepackage{url}
\usepackage{verbatim}
\usepackage{cite}
\usepackage{xcolor}
\usepackage{hyperref}
\usepackage{makecell}
\usepackage{graphicx}
\usepackage{multirow}
\usepackage{tabularx}
\usepackage{booktabs}
\usepackage{makecell}
\usepackage{soul}

\begin{document}

\title{Resilience Analysis Through Automated Stage Identification and Quantitative Assessment}

\author{Anna Strzoda, Krzysztof Grochla, Rafael Muñoz-Carpena%
\thanks{A. Strzoda and K. Grochla are with the Institute of Theoretical and Applied Informatics, Polish Academy of Sciences, Bałtycka 5, 44-100 Gliwice, Poland (e-mail: astrzoda@iitis.pl, kgrochla@iitis.pl). R. Muñoz-Carpena is with the Department of Agricultural and Biological Engineering, University of Florida, 1741 Museum Road, Gainesville, FL 32611-0570, USA (e-mail: carpena@ufl.edu).}%
\thanks{This research was funded in part by the National Science Centre, Poland, under the IMPRESS-U competition, project no. 2023/05/Y/ST7/00192, and by the U.S. National Science Foundation EAGER IMPRESS-U Award No. 2402580. For the purpose of Open Access, the author has applied a CC-BY public copyright license to any Author Accepted Manuscript (AAM) version arising from this submission. It was also partially funded by NATO Science for Peace and Security (SPS) Programme "Pathways for Infrastructure Resilience in Ukraine" (MYP G7979).}}

\markboth{IEEE Transactions on Reliability}%
{Strzoda \MakeLowercase{\textit{et al.}}: Recognition of Resilience Stages for Automatic Metric Evaluation}

\maketitle

\begin{abstract}
This paper introduced an ML-based approach, designed to be domain- and threat-agnostic, for automated recognition of PARA (Plan/Prepare, Absorb, Recover, Adapt) resilience stages in system-performance time series, enabling subsequent quantitative resilience analysis across different systems. We investigate four ML models for resilience stage recognition: Random Forest (RF), Support Vector Machine (SVM), Light Gradient Boosting Machine (LightGBM), and Extreme Gradient Boosting (XGBoost). We then use the ML models' predictions to quantify system resilience by computing corresponding resilience metrics from system-performance data observed before, during, and after disruptive events.
We evaluate the proposed approach using datasets from two distinct operational domain network systems and an airport transportation system, demonstrating its domain-agnostic applicability to resilience assessment.
The results demonstrate the effectiveness of resilience stage classification, with the best F1-scores ranging from 74.3\% to 99.4\% across all resilience stages.
The overall resilience metrics $R_t$ and $RL$ are estimated, with the best models' MAPE ranging from 0.35--0.82\% on IEEE 802.11 and LTE to 1.65--9.56\% for the airport.
These results provide a basis for threat-agnostic resilience analysis of complex engineering systems, enabling further assessment of how effectively a system withstands, absorbs, and recovers from disruptive events and stress conditions.
\end{abstract}


\section{Introduction}
\begin{figure}[h]
\includegraphics[width=\linewidth]{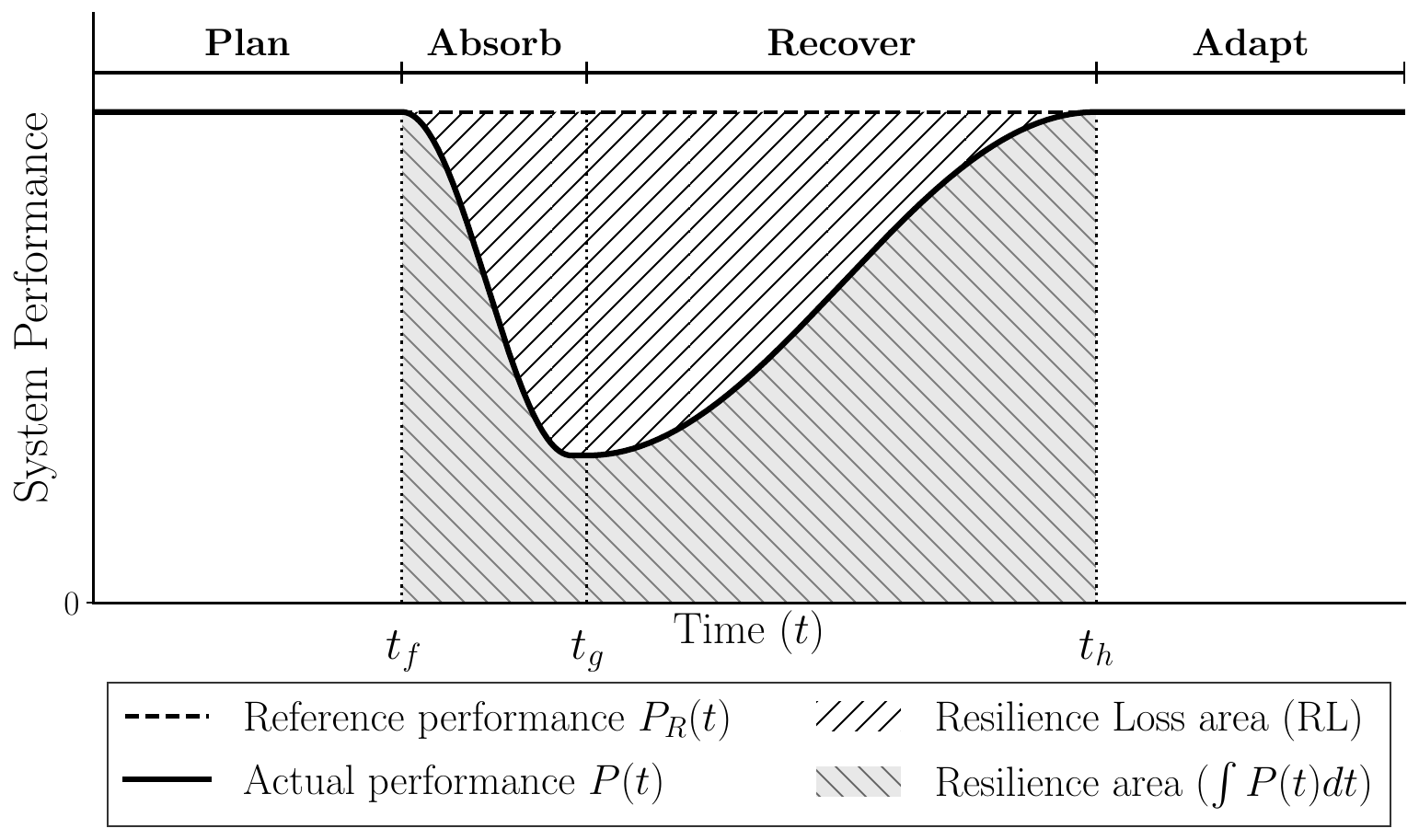}
\caption{Graphical illustration of the resilience curve archetype and the corresponding resilience stages.}
\label{fig:rt-metric}
\end{figure}

Resilience is a key property of complex systems and critical infrastructure, as it enables them to withstand disruptions, adapt to changing conditions, and maintain or recover an acceptable level of performance. In this context, resilience is a system's ability to prepare for threats and hazards, adapt to changing conditions, and recover from disruptions \cite{trump2025threat}.
Resilience has evolved into a widely recognized concept in both academic literature and professional practice, emphasizing how a system plans and prepares for, withstands and absorbs, recovers from, and adapts to various disruptions and threats \cite{linkov2019science, 10380780}. This sequence of capabilities is commonly referred to as the PARA (Plan/Prepare, Absorb, Recover, Adapt) model, and we use these stage names consistently throughout this study. The increasing complexity of modern technical systems has made resilience assessment an important aspect of reliability and safety analysis. In many applications, system performance over time follows a characteristic resilience curve archetype (Fig.~\ref{fig:rt-metric}), consisting of a stable pre-event stage (Plan), a disturbance stage in which performance degrades (Absorb), a gradual recovery process (Recover), and a post-recovery stage (Adapt). 

Focusing on system key performance indicators (KPIs), rather than the specific shock perturbing the system, provides a powerful, threat-agnostic approach to resilience analysis. This is critical in the context of modern compounded and unpredictable intentional and unintentional shocks affecting critical infrastructure systems today. In this framing, resilience evaluation is based on how deeply system performance degrades after the disturbance and how long it remains degraded. Accurate recognition of these stages can support fault detection, operator decision-making, maintenance planning, and dynamic reliability assessment~\cite{ganin2016operational}. 
Complex, interconnected systems, particularly critical infrastructure systems, require well-defined KPIs and quantitative resilience metrics to enable an objective and reproducible assessment of their resilience. However, the reliable computation of such metrics depends on accurately identifying the distinct stages of system performance evolution following a disturbance.

Automatic resilience stage recognition is a prerequisite for rigorous, scalable resilience evaluation because many established resilience metrics are defined relative to a specific stage, or a combination of stages, of the performance curve. The dominant conceptual models-the resilience triangle\cite{bruneau2003framework} and its generalization, the resilience trapezoid \cite{panteli2017}-partition a disruptive event into distinct stages (degradation, a degraded plateau, and restoration, corresponding broadly to the Absorb and Recover stages of PARA). The value of every summary metric depends on where the boundaries between these stages are placed. Resilience can be estimated as the integral of ``critical functionality''over a control interval, so both depth and duration of degradation must be properly delimited for the measure to be meaningful \cite{ganin2016operational, linkov2014changing}. The traditional approach to calculating these metrics is either to establish the stages from prior knowledge of when the disruptive event started and ended, or to tag them manually. However, when stage boundaries are drawn by hand, the resulting metrics become analyst-dependent, hard to precisely reproduce, and impractical to compute over large volumes of high-frequency data. Previous work~\cite{poulin2021infrastructure} has shown that performance-based metrics diverge substantially because the curve is operationalized inconsistently. Automating stage detection removes this subjectivity, enforces a consistent definition, and makes stage-resolved metrics computable at scale across many events and systems.

The need for precise resilience stage detection is evident in many metrics, which encode performance differences between stages. The $\Phi\Lambda\epsilon\Pi$ (FLEP) metric system from study\cite{panteli2017} is entirely stage-specific: $\Phi$ is the rate of performance decline during the disturbance (Absorb) stage, $\Lambda$ is the extent of degradation (the performance difference between the pre-event and degraded levels), $\epsilon$ is the duration spent in the degraded state, and $\Pi$ is the recovery rate during the restorative (Recover) stage, with a fifth ``area'' metric aggregating all four.

\begin{figure*}[t]
    \centering
    \includegraphics[width=.7\linewidth]{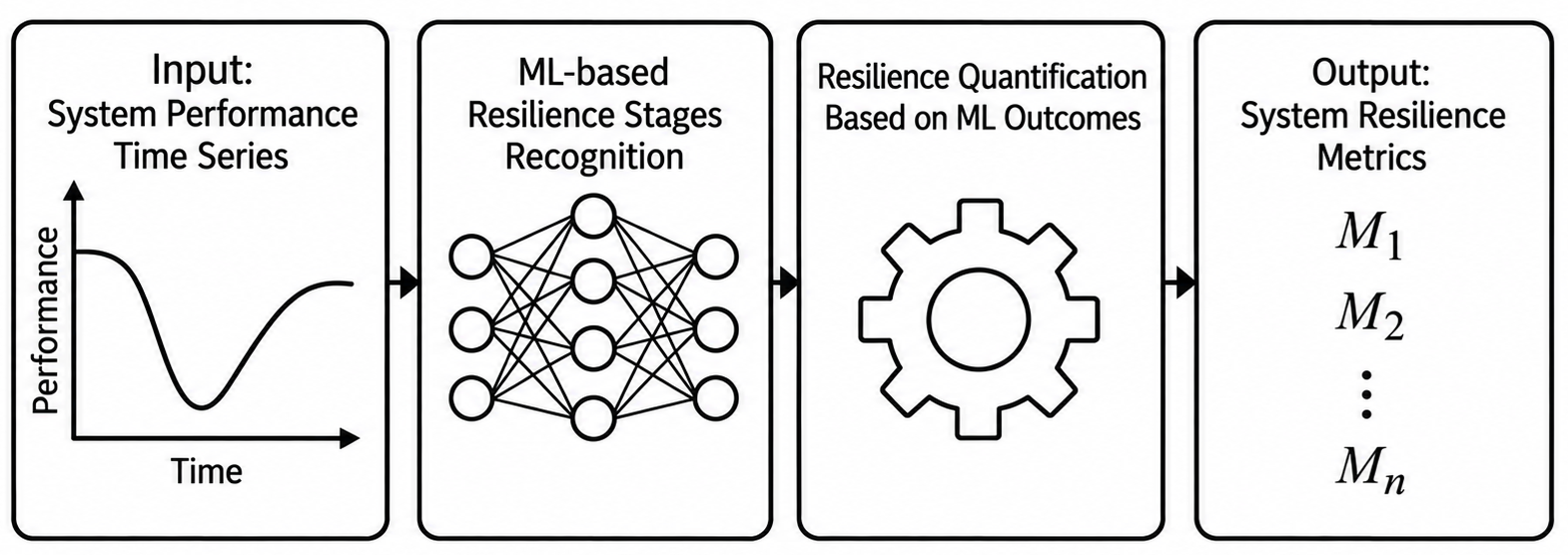}
    \caption{Workflow of ML-based resilience stage recognition for resilience quantification.}
    \label{fig:resilience_workflow}
\end{figure*}

In a survey \cite{poulin2021infrastructure}, the authors establish a taxonomy of resilience curve performance measures (availability, productivity, and quality) and summary metrics (magnitude, duration, integral, rate, threshold-based, and ensemble metrics), highlighting the importance of clearly defining the transition points between stages when comparing trajectories. The foundational pairing of robustness (the magnitude of performance loss at impact) and rapidity (the recovery slope) in \cite{bruneau2003framework} likewise captures differences localized to separate stages, while the integrated resilience-loss (``area under the curve'') metric is dominated by whichever stage is longest or deepest \cite{ganin2016operational}. Because each quantity changes with the chosen stage boundaries, reliable evaluation depends on identifying those boundaries automatically and consistently. To address this problem, we applied an ML-based approach for automatic resilience stage recognition in time-series performance data, to support broader analysis of system resilience and safety. This enables an automated pipeline from stage labels to stage-dependent resilience metric values, including quantification of the associated estimation errors. The main objectives of this study are as follows:
\begin{itemize}
    \item Develop an approach to automatically identify resilience stages from time-series measurements of system performance during disruptive events, complemented by a systematic comparative evaluation of multiple ML models across resilience stages.
    \item Develop an ML-driven resilience quantification approach in which the ML-identified resilience stages are directly used to compute quantitative resilience metrics, providing an integrated, multi-stage pipeline for automated resilience analysis from system performance time series (Figure \ref{fig:resilience_workflow}).
    \item Provide a publicly available benchmark dataset containing labeled resilience trajectories generated from comprehensive simulations of wireless communication systems under diverse disruption scenarios, enabling reproducible evaluation of resilience stage recognition and resilience analysis methods. The dataset contains data collected from IEEE 802.11 and LTE standards, covering both short- and long-range communication systems operating in stress scenarios.
    \item Conduct cross-domain validation by training the proposed models on wireless communication network data and evaluating their performance on an independent dataset from an airport transportation system, thereby assessing their generalization capabilities and domain-agnostic applicability to resilience stage recognition.
\end{itemize}
The paper is structured as follows: Section~\ref{sec:soa} reviews the state of the art in resilience stage recognition in system performance time series and resilience quantification. Section~\ref{sec:dataset} describes the dataset developed for training and testing the ML models, as well as an independent dataset from a different domain used for external validation. Section \ref{sec:methods} introduces our proposed ML-based approach for recognizing resilience stages and subsequently quantifying system resilience through resilience metrics calculated from the ML model outputs. In Section \ref{sec:results}, we evaluate the proposed ML models and apply them to system resilience analysis.
Finally, Section \ref{sec:conclusions} summarizes the main findings and presents the conclusions.

\section{Related work}
\label{sec:soa}
Quantification of system resilience is essential for ensuring objective and comparable assessments \cite{yang2023quantitative}. Accordingly, previous studies have proposed various resilience metrics reflecting different perspectives on resilience quantification \cite{sharma2018resilience,sharma2020regional,app13116452,LIU2021107868}. Their application has been widely investigated across diverse domains, including critical infrastructure, power systems, transportation systems \cite{BRUCKLER2024110176,RATHNAYAKA2025105271}, and various types of networks \cite{QI2024101882,ellison1997survivable}. The PARA (Plan, Absorb, Recover, Adapt) model provides a model for assessing how effectively a system prepares for, withstands, absorbs, and recovers from disruptive events, and adapts to changing conditions. Within this model, the computation of stage-dependent resilience metrics from system performance time series requires the identification of PARA stages, their transition points, and corresponding temporal segments of the resilience curve \cite{linkov2019science,panteli2017,de2025quantitative}. Automated PARA stage classification is therefore a fundamental step in computing stage-dependent resilience metrics from system performance time series. This makes automated PARA stage identification particularly relevant to the automation of quantitative resilience analysis.

Since the PARA stages are reflected in the temporal evolution of system performance, their automated identification can be formulated as a time series analysis problem. Methods for analyzing such data have been extensively investigated in the context of system monitoring, fault diagnosis, and operating state identification.
Previous studies have reviewed conventional and deep learning methods for time series classification \cite{abanda2019review,faouzi2022time,ismail2019deep}. Related research directions include failure prognosis, which focuses on estimating the remaining useful life or time to failure of engineering systems \cite{8842589}, and change point detection \cite{aminikhanghahi2017survey}.

Classical statistical approaches, such as autoregressive integrated moving average (ARIMA) models \cite{box1970time}, offer computational efficiency, interpretability, and applicability to relatively small datasets. However, their predominantly linear structure limits their ability to represent complex nonlinear performance dynamics, and they are not directly designed for semantic labeling of individual observations across heterogeneous time series. In contrast, ML methods can learn nonlinear patterns from multiple time series, although they introduce challenges related to overfitting and reduced interpretability. Hybrid approaches combining statistical and data-driven reasoning are also emerging \cite{runge2023causal}. The ability of ML methods to capture complex performance patterns across different disruption scenarios makes them particularly relevant to automated resilience stage classification.

Closely related research directions are time series segmentation and sequential state identification. Change Point Detection (CPD) aims to identify time instants at which the statistical properties of a time series change significantly and is widely applied to signal segmentation, fault detection, and activity recognition \cite{6682893,10292634}. Software libraries such as \textit{ruptures} \cite{TRUONG2020107299} provide implementations of various CPD algorithms, whereas \textit{STUMPY} \cite{law2019stumpy} supports shape-based time series segmentation. Hidden Markov Models, implemented, for example, in the \textit{hmmlearn} library, provide an alternative approach to modeling processes characterized by sequences of latent states. Although these methods can identify segment boundaries or underlying states, they do not inherently associate them with predefined resilience stages. Their application to PARA stage recognition therefore requires an additional semantic interpretation or classification mechanism.

Despite the growing application of ML to infrastructure resilience analysis and enhancement \cite{11181573}, automated classification of individual PARA resilience stages remains largely unexplored. The study \cite{poulin2021infrastructure} formalized the resilience curve by distinguishing characteristic events and transitions associated with disruption, cascading failures, and recovery. However, existing ML-based studies have primarily focused on identifying resilience archetypes rather than recognizing individual resilience stages. Resilience archetypes represent characteristic trajectories of system performance throughout disruption and recovery. For example, \cite{hsu2025dissecting} characterizes resilience curves using hand-crafted features describing magnitude, duration, integral, recovery rate, and stability, followed by k-means clustering to identify representative archetypes. Similarly, \cite{li2024unraveling} combines Dynamic Time Warping (DTW) with k-means clustering to group resilience curves by shape similarity.

While these approaches enable the identification of recurring resilience patterns, they operate at the level of entire resilience curves and do not directly address the classification of individual PARA stages or the identification of their temporal boundaries. In contrast, point-wise resilience stage classification corresponds to the sequence-labeling setting, in which each observation is assigned a class label rather than assigning a single label to the entire time series \cite{nakata2014modified}. This distinction is particularly important for quantitative resilience assessment, as the identification of an overall resilience archetype does not necessarily provide the temporal information required to compute stage-dependent resilience metrics.

A further limitation is the scarcity of publicly available datasets capturing the full temporal evolution of engineering system performance under hazardous or disruptive events, including pre-event, disruption, and post-event recovery periods. In particular, both real-world and simulation-based datasets in which individual system performance observations are annotated with the corresponding PARA resilience stages remain limited. The lack of such datasets restricts the development, training, validation, and systematic comparison of ML-based resilience stage classification methods across different engineering domains and disruption scenarios.

Beyond these methodological and data-related limitations, existing research has given limited attention to comprehensive, domain- and threat-agnostic frameworks integrating automated PARA stage classification with subsequent resilience quantification. Although numerous methods and metrics have been developed for individual aspects of resilience assessment, the integration of these tasks within a unified workflow remains insufficiently explored. Since the computation of stage-dependent resilience metrics relies on identifying the corresponding PARA stages and their temporal boundaries, addressing these tasks independently limits the potential for fully automated quantitative resilience assessment.

To address these research gaps, this study proposes a comprehensive, domain- and threat-agnostic framework integrating automated PARA resilience stage classification with subsequent resilience quantification. Given the nonlinear and heterogeneous nature of system performance trajectories, resilience stage recognition is formulated as a supervised multiclass classification problem at the individual observation level. The identified stages and their temporal boundaries provide the basis for computing resilience metrics within the same analytical workflow. By integrating these tasks, the proposed framework enables automated, multi-stage resilience analysis and quantitative assessment of how effectively engineering systems withstand, absorb, and recover from disruptive events.

\section{Dataset}
\label{sec:dataset}

\subsection{Network systems - IEEE 802.11 and LTE datasets}
\label{sec:wireless-dataset}
This study analyzes simulated data generated using the OMNeT++ discrete-event simulation framework~\cite{omnetpp} to evaluate wireless network performance under stress scenarios involving disruptive events. The simulations cover both IEEE 802.11 and LTE networks, representing wireless local area and cellular wide-area communication technologies, respectively. The resulting dataset consists of time series of packet delivery probability, which indicates network performance. The dataset is publicly available on Zenodo~\cite{strzoda2026recognition}. To the best of our knowledge, it is the first dataset of its kind reported in the literature, enabling benchmarking of methods for resilience stage analysis.

The dataset is provided in CSV format, with each file containing a single time series of packet delivery probability representing network performance under a specific stress scenario. Each stress scenario corresponds to a disruptive event simulated in OMNeT++, involving the failure of critical network infrastructure components, namely evolved NodeBs (eNodeBs) in LTE networks or wireless access points (APs) in IEEE 802.11 networks, resulting in a temporary degradation of network performance. For both LTE and IEEE 802.11 networks, the disruptive event is modeled as disabling one, two, or three of the four available eNodeBs or APs, respectively. These failures disrupt wireless connectivity and may prevent network hosts from successfully transmitting packets, reducing packet delivery probability. Consequently, the dataset captures the impact of infrastructure outages on packet delivery performance in both LTE and IEEE 802.11 networks.

In total, the dataset contains 457 time series, including  280  of IEEE 802.11  and 177 of LTE. Each time series corresponds to a single simulation run, since the data are simulated. 
Each simulation scenario covers $\Delta$ seconds of network operation. In this study, we consider $\ Delta=200$ seconds. During the simulation run, a disruptive event occurs at time $t_f$ and lasts for $\delta=t_g-t_f$, where $t_g \ge t_f$ is the end time of the disruptive event. The simulation parameters $t_f=40$, $t_g \in \{60, 80\}$ [s].  Each series reflects the characteristic shape of a resilience curve and its corresponding resilience stages, as shown in Figure \ref{fig:rt-metric}. Resilience is commonly described as a sequence of stages \cite{linkov2019science}: Plan, Absorb, Recover, and Adapt.

Regarding the Adapt stage, the data collected for this study show that system performance returns to its (pre-disturbance) level and remains stable over time. This behavior is consistent with a resilient system, where functionality is restored to the level observed before the disruptive event. Consequently, the post-recovery observations exhibit the same performance characteristics as in the Plan stage. Therefore, for this analysis, we incorporate the post-recovery adaptation stage into the Plan class. The dataset is labeled with the corresponding resilience stages.

The network dataset, consisting of 457 CSV files collected from IEEE 802.11 and LTE network simulations, was randomly divided into training and test sets using an 80:20 split. The training set contained 365 files (80\%), while the remaining 92 files (20\%) were reserved for testing. The split was performed at the file level, ensuring that each simulation scenario was assigned exclusively to either the training or test set. The test set was used to evaluate the performance of the trained machine learning models on previously unseen network scenarios.

\subsection{Airport dataset}
\label{sec:airport-dataset}
For external validation, we use a dataset from an airport transportation system, a domain different from wireless networks. The dataset was generated using an airport operations simulation model representing one of the airports in the United States. It is used exclusively to evaluate the proposed ML models trained on IEEE 802.11 and LTE data (Section \ref{sec:ltewifi}), without additional training or fine-tuning, allowing us to assess their cross-domain generalization in an entirely unseen operational environment.

The Airport dataset captures aircraft arrival and departure queue dynamics under disruptive events. The disruptive event is modeled as a severe weather disturbance that significantly affects airport operations. Under normal conditions, the airport operates efficiently, maintaining stable traffic flow and minimal congestion. The onset of the disturbance substantially reduces runway capacity, increases flight cancellations, and causes arrival and departure queues to build, resulting in operational disruptions. Once the adverse weather conditions subside, normal operating capacity is restored. However, the airport requires additional time to clear the accumulated backlog and gradually return to its pre-disruption performance levels. Weather disturbances are simulated by applying weather-dependent factors that modify runway capacity and flight cancellation probability. Severe weather conditions reduce operational capacity and increase the likelihood of flight cancellations. Once normal weather conditions are restored, these parameters return to their baseline values.

The dataset consists of 270 time series stored in CSV format, each representing an independent simulation run involving a disruptive event. System performance is characterized by the total number of aircraft queued for arrivals and departures. Because queue length is inversely related to system resilience, we transform this metric into an inverted performance signal, denoted by $\rho$, such that an increase in queue length corresponds to a decrease in system performance, and vice versa. Each simulation spans 200 discrete time steps, with no predefined physical time unit.

The simulation scenarios consider different disruption onset times, durations, and random seeds. Specifically, disruptive events begin at simulation steps $t_f \in \{50,60,70\}$ (Start)  and last for $\delta\in \{70,80,80\}$. For each of the nine combinations of disruption onset and duration, 30 independent simulations were performed, resulting in 270 simulation scenarios.

Each time series captures how system performance evolves across resilience stages. As post-recovery observations exhibit performance characteristics similar to those of the Plan stage, the Adapt stage is merged with the Plan class, consistent with the approach adopted for the IEEE 802.11 and LTE datasets. Consequently, three classes are considered for resilience stage recognition: Plan, Absorb, and Recover. The dataset is labeled with the corresponding resilience stages.

\begin{figure*}[ht]
\includegraphics[width=\linewidth]{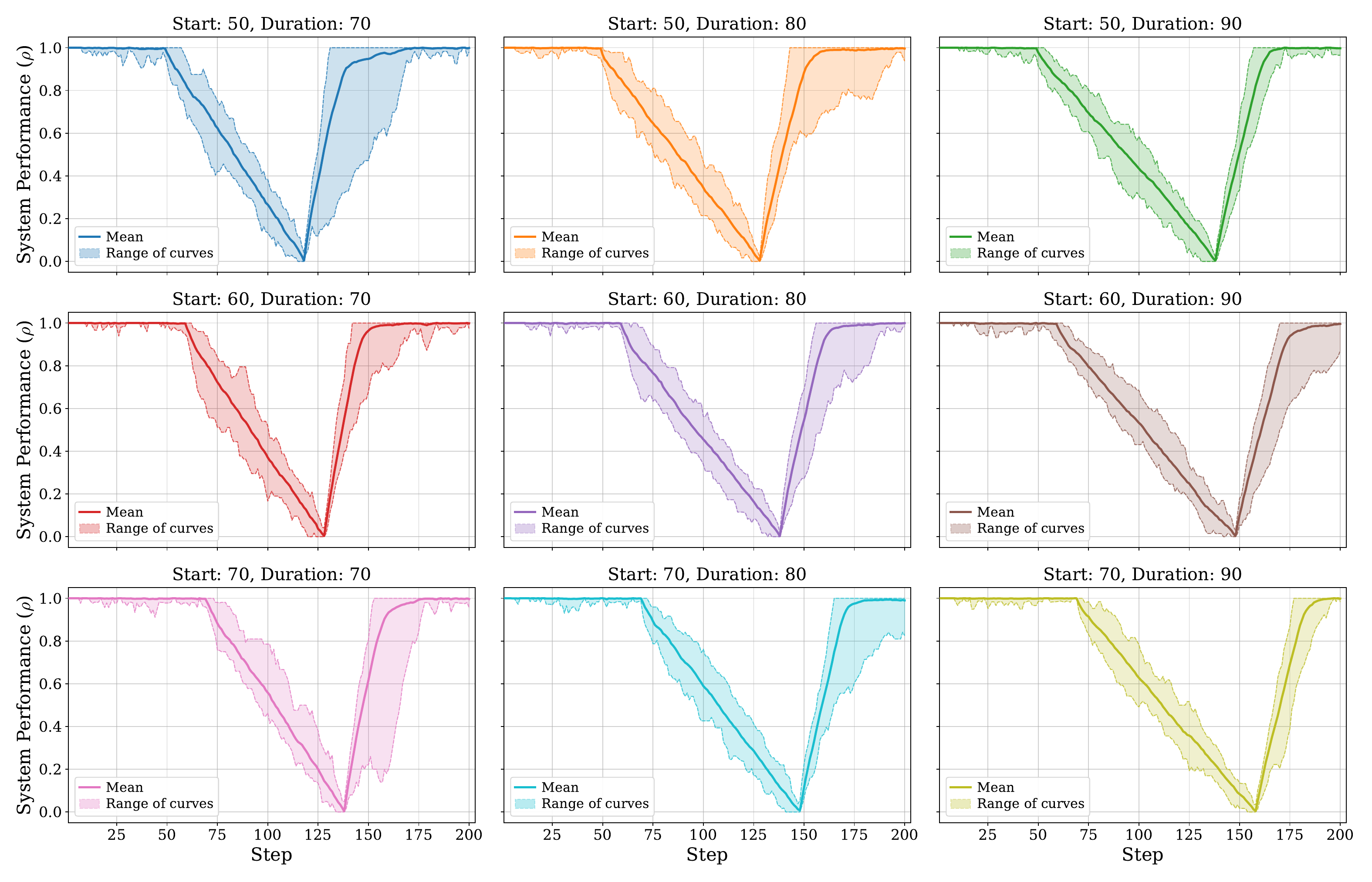}
\caption{System performance ($\rho$) trajectories from the Airport simulation dataset used for external validation, generated under different parameter configurations ($t_f \in \{50, 60, 70\}$, $\delta \in \{70, 80, 90\}$). Each subplot presents the mean trajectory, with the shaded area indicating the range of 30 trajectories generated using different random seeds for each parameter configuration.}
\label{fig:airport-data}
\end{figure*}

Figure \ref{fig:airport-data} shows aggregated system performance trajectories across nine parameter configurations, where $t_f$ denotes the disruption start step (``Start'' in Fig.\ref{fig:airport-data}) and $\delta$ denotes the disruption duration (``Duration'' in Fig.\ref{fig:airport-data}).  Each subplot summarizes a subset of 30 independent simulation runs for each parameter configuration, e.g., $t_f=50$ and $\delta=70$. 

To construct the summary representation for a given parameter set, we aggregate the curves pointwise at each discrete simulation step. Specifically, at each step, the thick solid line represents the arithmetic mean of the 30 trajectories for that configuration, capturing the typical progression of the degradation and recovery stages. Around this mean trajectory, the lower and upper dashed contours of the shaded envelope are determined by taking the minimum and maximum performance values recorded across all 30 curves at that exact time step. Consequently, the vertical width of the shaded band at any point along the horizontal axis directly reflects the full empirical spread and stochastic variability of the generated curves for a given set of simulation parameters.

\section{Methods}
\label{sec:methods}
Figure \ref{fig:resilience_workflow} provides a schematic overview of the proposed ML-based system for resilience-stage recognition in system-performance time series and subsequent resilience analysis.
We investigate ML-based approaches to automatically identify resilience stages within the standard resilience-curve archetype, which describes system behavior before, during, and after a disruptive event that leads to performance degradation. The identified stages are subsequently used to assess system resilience by quantitatively characterizing their properties using a set of resilience metrics.
\begin{table}
\begin{center}
\caption{Symbols and notation used in this study}
\label{tab:nomenclature}
\begin{tabular}{| c | c |}
\hline
Symbol & Description \\
\hline
$\Delta$ & Duration of a single simulation run\\
\hline
$t_f$ & Disruption onset time\\
\hline
$t_g$ & Recovery onset time\\
\hline
$\delta$& Shock duration, $\delta=t_g-t_f$\\
\hline
$t_h$ & Time of Recovery completion\\
\hline
RL & Resilience loss area\\
\hline
$R_t$ & Overall resilience\\
\hline
$\epsilon$ & The absorb stage duration\\
\hline
$\Lambda$ & Magnitude of resilience drop\\
\hline
$P(t)$ & actual system performance as a function of time\\
\hline
$P_R(t)$ & reference system performance as a function of time\\
\hline
$\rho$ & Airport system performance metric\\
\hline
$X$ & Expected system performance\\
\hline
$l$ & Sliding window size\\
\hline
$\mathcal{X}$ & The collected time series dataset\\
\hline
$N$ &  The cardinality of the set of time series $\mathcal{X}$\\
\hline
$\mathcal{Y}$ & set of ground truth label sequences\\
\hline
$x^{(i)}$ & \makecell[l]{$i$-th time series of system\\performance metric} \\
\hline
$x^{(i,t)}$ & $t$-th sliding window of the $i$-th time series \\
\hline
$y^{(i)}$ & $i$-th sequence of resilience stages labels\\
\hline
$y^{(i,t)}$ & label for $t$-th sliding window in $i$-th time series\\
\hline
$x_b^{(i)}$ & baseline time series\\
\hline
$n$ & Median filter window size \\
\hline
$W_j^{(i)}$ & the $j$-th local window of the $i$-th time series\\
\hline
\end{tabular}
\end{center}
\end{table}

The proposed approach aims to provide a domain-agnostic system for threat-agnostic resilience analysis. By focusing on universal patterns in system behavior and the underlying mathematical structure of resilience trajectories, rather than domain-specific rules, the methodology can extend to a broader range of application areas by retraining on more diverse datasets. Furthermore, if the performance time series are normalized (e.g., with respect to their pre-event baseline), the approach can be applied consistently across systems with different metric magnitudes and units.\begin{figure}[htbp]
    \centering
    \includegraphics[width=.85\linewidth]{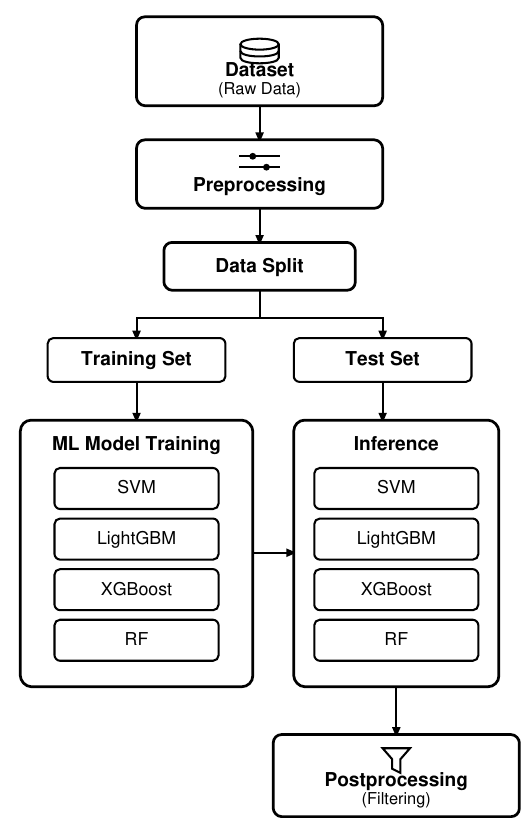}
    \caption{Schematic overview of the ML-based resilience stage recognition development. The pipeline includes data preprocessing, training of ML models (SVM, LightGBM, XGBoost, and RF), inference, and post-processing.}
    \label{fig:graphical-abstract}
\end{figure}

Table~\ref{tab:nomenclature} lists the symbols and notation used throughout this study.\subsection{ML models}
We employ four well-established and widely used supervised classification algorithms for resilience stage recognition in system-performance time series: Random Forest (RF) \cite{hastie2008random}, Support Vector Machines (SVM) \cite{bishop2006pattern_svc}, Extreme Gradient Boosting (XGBoost) \cite{chen2016xgboost}, and Light Gradient Boosting Machine (LightGBM) \cite{ke2017lightgbm}. These methods are commonly used for classification tasks and represent distinct learning paradigms. RF provides an ensemble of randomized decision trees and is well suited to nonlinear classification problems involving complex interactions among input features. SVM provides a different, margin-based classification approach and can model nonlinear decision boundaries through kernel functions. XGBoost and LightGBM are gradient-boosted decision tree methods that construct predictive models by sequentially combining weak tree learners while incorporating computational and algorithmic improvements for efficient learning \cite{chen2016xgboost, ke2017lightgbm}. Including methods from the bagging, margin-based learning, and gradient boosting families allows the resilience stage classification task to be evaluated across different supervised learning paradigms.

Figure \ref{fig:graphical-abstract} illustrates the methodological workflow adopted to develop the proposed ML-based approach for resilience stage recognition. The workflow comprises dataset preprocessing, partitioning the dataset into training and test sets, training the ML models, applying the trained models to classify unseen data, and post-processing the resulting model outputs.

\subsubsection{Support Vector Machines}
The Support Vector Classifier (SVC) assigns a class label to an input sample represented by a vector of extracted features. Its core objective is to find an optimal separating hyperplane that maximizes the margin - the distance between the decision boundary and the closest training samples from each class, known as support vectors.

To handle complex, non-linearly separable data, SVC implicitly maps the input features into a higher-dimensional feature space where linear separation becomes feasible. Furthermore, the classifier employs a soft-margin approach controlled by a regularization parameter, which balances the trade-off between maximizing the separating margin and tolerating a certain degree of misclassification to prevent overfitting.

Since standard SVC is inherently a binary classifier, multiclass classification is handled using a one-vs-one strategy. In this approach, a separate binary classifier is trained for every unique pair of classes. When predicting the label for a new, unseen sample, all pairwise classifiers are evaluated, and the final class assignment is determined by combining their individual decisions (typically via majority voting).

\subsubsection{Light Gradient Boosting Machine}
LightGBM follows the same general gradient boosting principle as XGBoost but differs mainly in the way individual trees are constructed. LightGBM uses a leaf-wise growth strategy, in which the leaf yielding the largest reduction in the loss function is split at each step. This approach can improve computational efficiency and predictive performance, particularly for large datasets.

\subsubsection{Random Forest}
The Random Forest (RF) classifier is an ensemble learning method that combines the predictions of multiple decision trees to improve generalization and reduce overfitting. Each tree in the forest is trained independently on a bootstrap sample (a random subset drawn with replacement) of the training data. Additionally, at each node split, only a randomly selected subset of input features is considered, which decorrelates the individual trees. For multiclass classification, a new sample is evaluated by all trees in the ensemble, and the final class label is assigned by aggregating their predictions via majority voting.

\subsection{Training and Test Sets}
Let $\mathcal{X} = \{ x^{(i)} \}_{i=1}^{N}$ denote a set of time series, where each time series is defined as
\begin{equation}
x^{(i)} = \{x^{(i)}_1, x^{(i)}_2, \dots, x^{(i)}_{T}\},
\end{equation}  and $N=457$ is the total number of time series in the dataset described in Section \ref{sec:wireless-dataset} used for ML model training. Parameter $T$ denotes the length of each time series. Let 
\begin{equation}
    \mathcal{Y} = \{ y^{(i)} \}_{i=1}^{N}
\end{equation} denotes the corresponding set of ground truth label sequences, where
\begin{equation}
y^{(i)} = \{y^{(i)}_1, y^{(i)}_2, \dots, y^{(i)}_{T}\}.
\end{equation}

The training set is constructed using a sliding window approach. For a fixed window size $l \in \mathbb{N}$, each feature vector for the $i$-th time series is defined as:
\begin{equation}
\mathbf{x}^{(i,t)} = [x^{(i)}_{t-l+1}, x^{(i)}_{t-l+2}, \dots, x^{(i)}_t], \quad t = l, \dots, T.
\end{equation}

The corresponding label for each window is given by:
\begin{equation}
y^{(i,t)} = y^{(i)}_t.
\end{equation}

Thus, the training set consists of pairs:
\begin{equation}
\{(\mathbf{x}^{(i,t)}, y^{(i,t)})\}_{i=1,\dots,N;\; t=l,\dots,T}.
\end{equation}
In this study, we set $l = 20$, chosen empirically.
Each time series observation is classified using its corresponding feature vector together with information derived from preceding historical observations.
The proposed feature representation is designed for near-online deployment, as classifying each observation relies exclusively on the current measurement and $l-1$ historical observations, together with the median filtering described in Section~\ref{sec:postproc}. Because the centered median filter uses the $m$ predictions that follow the current one, the method introduces a fixed delay of $m$ samples but requires no further future information, enabling near-real-time inference.\subsection{Postprocessing}\label{sec:postproc}
As part of the postprocessing stage, median filtering is applied to reduce short-term fluctuations and eliminate sporadic classification errors. A centered window $W_j = \{\hat{y}_{j-m}, \dots, \hat{y}_j, \dots, \hat{y}_{j+m}\}$ of size $n = 2m+1$ is used (with $n=5$, $m=2$), replacing each prediction with the median of its local neighborhood, effectively suppressing impulsive noise while preserving true stage transitions. This approach is widely recognized for its robustness to outliers and edge-preserving characteristics in signal processing \cite{tukey1977exploratory, gallagher1981median}. For near-online implementation, this centered filter introduces a fixed decision latency of $m=2$ time steps; alternatively, a causal trailing median filter can be employed when strictly zero latency is required.

Let 
\begin{equation}
\mathcal{\hat{Y}}=\{\hat{y}^{(i)}\}_{i=1}^N
\end{equation}
\begin{equation}
    \hat{y}^{(i)} = \{\hat{y}^{(i)}_{l}, \hat{y}^{(i)}_{l+1}, \cdots, \hat{y}^{(i)}_{T} \}
\end{equation}
where $\hat{y}^{(i)}$ denotes a sequence of outputs of the ML model, where each $\hat{y}^{(i)} \in \mathcal{\hat{Y}}$, corresponds to the predicted class label assigned by the model to the observation at a step $t$ in the $i$-th time series. 
A median filter with an odd window size $n = 2m+1$ is applied as follows.For each element $\hat{y}^{(i)}_j \in \hat{y}^{(i)}$, where $j \in \{l,\dots,T\}$, a local window corresponding to the element is defined as:
\begin{equation}
W_j^{(i)} = \{\hat{y}^{(i)}_{j-m}, \dots, \hat{y}^{(i)}_{j+m}\}
\end{equation}

The filtered value is then given by:
\begin{equation}
\tilde{y}^{(i)}_j = \operatorname{median}(W_j^{(i)})
\end{equation}

Finally, the post-processed ML model-based result for a single time series $x^{(i)}$ is defined as follows
\begin{equation}
    \tilde{y}^{(i)}=\{\tilde{y}^{(i)}_l, \tilde{y}^{(i)}_{l+1},...,\tilde{y}^{(i)}_T\}
\end{equation}

\subsection{Resilience Quantification}
One of the main goals of resilience assessment is to evaluate a system's capacity to withstand and respond to disruptions \cite{yang2023quantitative}. To support this assessment, resilience metrics provide quantitative measures of performance degradation and recovery, enabling the characterization and parameterization of system performance throughout the disruption and recovery process \cite{panteli2017, bruneau2003framework, de2025quantitative}. 
Resilience metrics support the analysis, modeling, and comparison of systems and facilitate the development of more adaptive and robust infrastructures. Quantitative methods employ mathematical and statistical models to describe, measure, and evaluate resilience processes objectively.

The literature has proposed numerous resilience metrics \cite{sharma2018resilience, sharma2020regional, app13116452, LIU2021107868}. In this study, we consider resilience metrics that characterize both the system's overall resilience and a selected stage of the resilience process. Table \ref{tab:resilience_metrics} summarizes the resilience metrics investigated in this study. All considered resilience metrics are domain-agnostic and can therefore be applied to a wide range of engineering systems.

\begin{table*}[t]
\caption{Resilience metrics considered in this study}
\label{tab:resilience_metrics}
\centering
\renewcommand{\arraystretch}{1.4}
\begin{tabularx}{\textwidth}{
  >{\centering\arraybackslash}X |
  >{\centering\arraybackslash}X |
  >{\centering\arraybackslash}X |
  >{\centering\arraybackslash}X
}
\hline
\textbf{Symbol} & \textbf{Description} & \textbf{Measuring Unit} & \textbf{Mathematical Expression}\\
\hline
RL  \cite{bruneau2003framework,panteli2017}
& Resilience Loss area (see Fig.~\ref{fig:rt-metric})
& Performance $\times$ Time
& Eq.~(\ref{eq:RL})\\
\hline
$R_t$ \cite{de2025quantitative,LIU2021107868}
& Overall system resilience - ratio of actual to expected area under the resilience curve
& Dimensionless (0--1)
& Eq.~(\ref{eq: R_t})\\
\hline
$\epsilon$ \cite{app13116452,panteli2017}
& Duration of the degraded state
& Time (or steps)
& Eq.~(\ref{eq:epsilon})\\
\hline
$\Lambda$ \cite{app13116452,panteli2017}
& Resilience degradation
& Dependent on system performance units
& Eq.~(\ref{eq:lambda})\\
\hline
\end{tabularx}
\end{table*}

The resilience metrics considered in this study are described below.\begin{itemize}
\item Metric $\epsilon$ from the $\Phi\Lambda\epsilon\Pi$ system\cite{app13116452, panteli2017}. It describes how long the degraded post-disturbance state lasts, measured as the time interval between the onset of the Absorb stage and the moment the system enters the Recover stage:\begin{equation}
    \label{eq:epsilon}
    \epsilon = t_g-t_f
\end{equation}
where $t_g$ is the time the system enters the recovery stage, and $t_f$ is the moment the disruptive event occurs, as shown in Figure \ref{fig:rt-metric}. The unit of measurement for $\epsilon$ is time.

\item Metric $R_t$ \cite{de2025quantitative, LIU2021107868} - the overall level of resilience of the system is defined as the ratio of the area under the actual performance of the system $P(t)$ to the area under the reference performance curve $P_R(t)$ over the time interval considered, $t_f \le t\le t_h$, with $t_h\ge t_f$, as shown in Figure \ref{fig:rt-metric}. The overall system resilience $R_t$ is defined as follows

\begin{equation}
R_t =\frac{\displaystyle \int_{t_f}^{t_h} P(t)\,dt}
{\displaystyle \int_{t_f}^{t_h} P_R(t)\,dt},
\label{eq: R_t}
\end{equation}
where $P(t)$ is the actual system performance curve, and $P_R(t)$ represents the reference performance, corresponding to the case where no disruptive event occurs and the system maintains its nominal performance level throughout the considered time period. The time interval $[t_f, t_h]$ is the time between the start of the disruptive event ($t_f$) and complete restoration ($t_h$).
A value of $R_t$ close to 1 indicates high system resilience and an effective ability to maintain performance despite disruptions. In contrast, a value close to 0 reflects low resilience and a limited capacity to preserve functionality under adverse conditions.

The resilience metric $R_t$ (Eq. \ref{eq: R_t}) was individually evaluated for each time series $x^{(i)}$ in the data set by representing the performance function $P(t)$ with the corresponding series $x^{(i)}$ and the reference performance function $P_R(t)$ with the corresponding baseline time series $x_{b}^{(i)}$.\item The Resilience Loss (RL) metric \cite{app13116452, bruneau2003framework, panteli2017} provides a global quantitative measure of system resilience. Following the formulation in \cite{de2025quantitative, LIU2021107868}, RL represents the cumulative deviation of the actual system performance $P(t)$ from the reference performance $P_R(t)$ over the analyzed time horizon, and is therefore geometrically interpreted as the area enclosed between the two performance trajectories.

For a single disruption event in the $i$-th time series, bounded by the ML-predicted Absorb onset $t_f$ and Recover completion $t_h$, RL is defined as follows:\begin{equation}
\label{eq:RL}
RL = \int_{t_f}^{t_h}\left[ P_R(t) - P(t)\right] dt
\end{equation}

\item Metric $\Lambda$ from the $\phi \epsilon \Lambda \Pi$ system \cite{panteli2017} quantifies system degradation as the difference between the expected system performance (baseline) and the actual performance during the Absorb stage. The expected performance can be specified by a domain expert based on knowledge of the system's behavior under normal operating conditions. Alternatively, it can be estimated from the model results by calculating the average system performance during the Plan stage. However, this approach may introduce additional estimation errors. Therefore, the choice of method for determining the expected performance is left to the analyst's discretion, depending on the available data and expert knowledge.

\begin{equation}
\mathcal{T}^{(i)}_{\text{Absorb}} =
\left\{ t \in \{1, 2, \ldots, T\} \;\middle|\; \tilde{y}^{(i)}_t = \text{Absorb} \right\}.
\end{equation}

The resilience metric $\Lambda$ for the $i$-th time series $\mathbf{x}^{(i)} = \{x^{(i)}_1, x^{(i)}_2, \ldots, x^{(i)}_T\}$ is then defined as

\begin{equation}
\label{eq:lambda}
\Lambda\big(\mathbf{x}^{(i)}\big)
=
X -
\frac{1}{|\mathcal{T}^{(i)}_{\text{Absorb}}|}
\sum_{t \in \mathcal{T}^{(i)}_{\text{Absorb}}}
x^{(i)}_t,
\end{equation}
where $X \in \mathbb{R}$ is the expected system performance.

\end{itemize}

\section{Results}
\label{sec:results}
\subsection{Resilience Stages Recognition}
\begin{figure}[h] 
     \subfloat[LTE (sample from the test set)\label{fig:curvelte}]{
         \includegraphics[width=\linewidth]{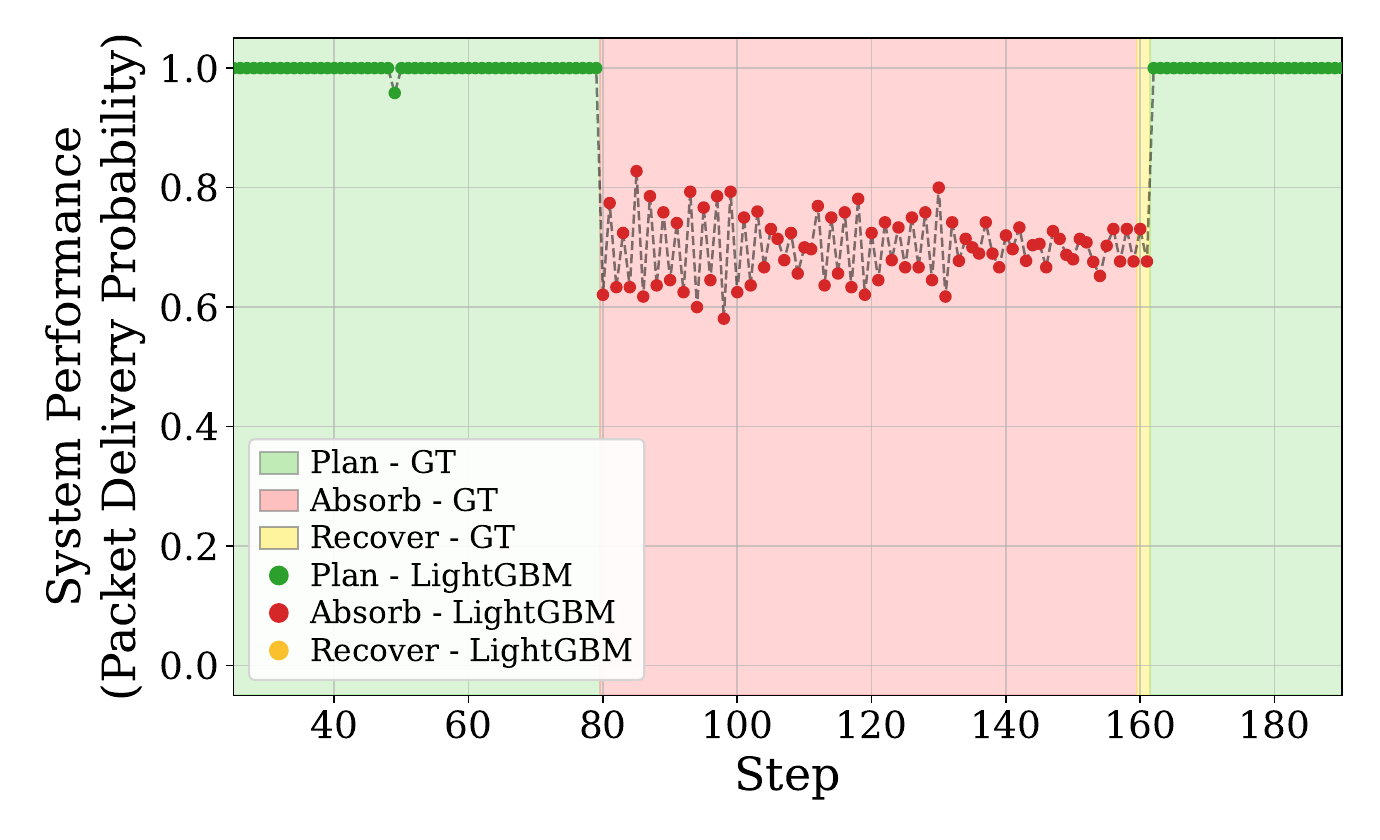}
     }\\
     \subfloat[IEEE 802.11 (sample from the test set) \label{fig:curvewifi}]{
         \includegraphics[width=\linewidth]{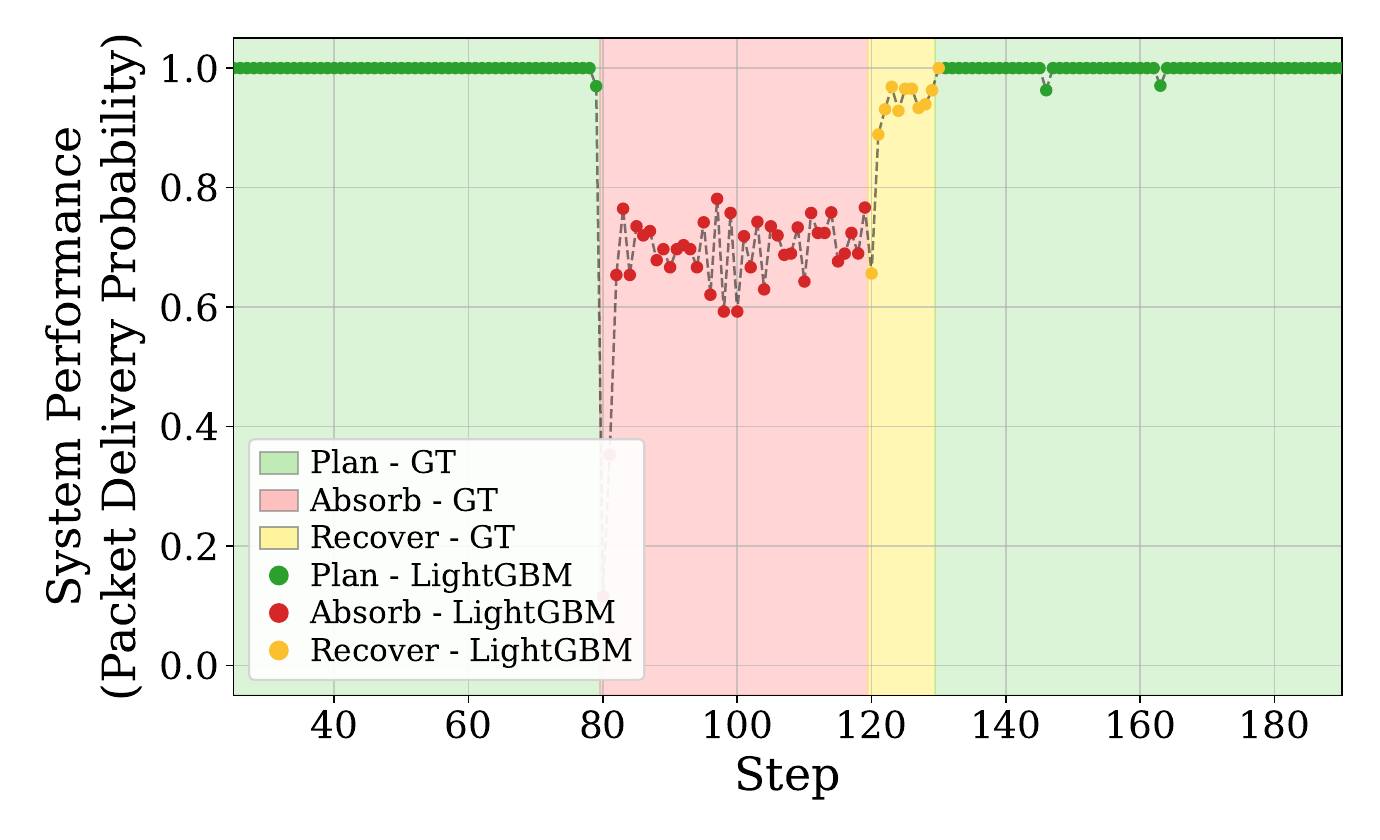}
     } \\
     \subfloat[Airport (sample from the external validation set)\label{fig:curveairport}]{
         \includegraphics[width=\linewidth]{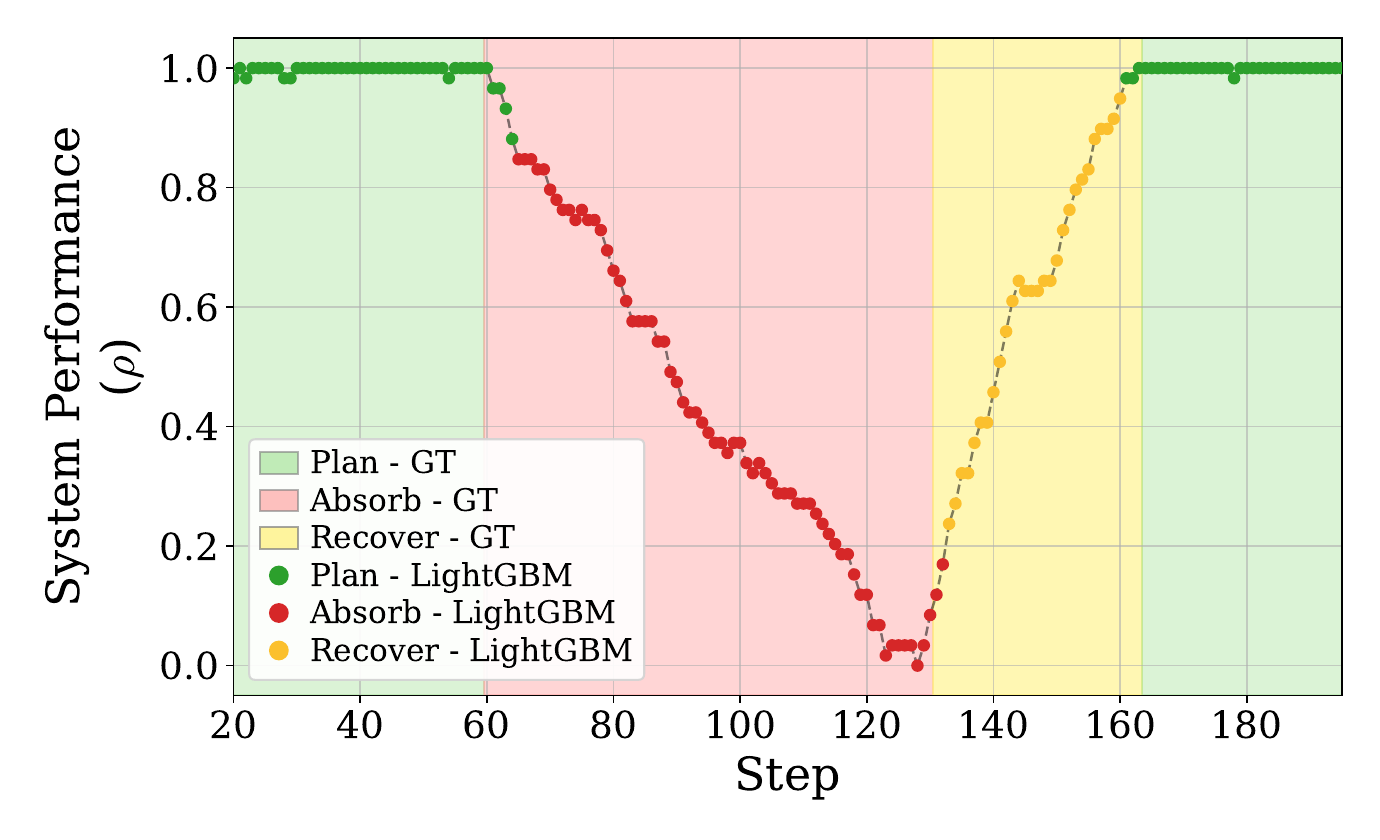}
     }
\caption{System-performance time series from the dataset described in Section~\ref{sec:dataset}, illustrating representative curves from IEEE 802.11, LTE, and Airport systems, together with the corresponding resilience-stage classifications obtained using the LightGBM-based approach vs. ground truth (GT).}
\label{fig:curves}
\end{figure}

Figure \ref{fig:curves} presents representative system-performance time series for IEEE 802.11, LTE, and Airport systems from datasets described in Section~\ref{sec:dataset}. The plots are zoomed in to highlight the most relevant portions of the time series. Although the curves show different temporal characteristics and performance degradation patterns, they all follow the Plan, Absorb, and Recovery stages in the evolution of system performance.
The background colors represent the ground-truth resilience stages, whereas the colors of the individual data points indicate the stages classified by the one of the ML models. For illustration purposes, the results obtained using LightGBM are presented as a representative example. Agreement between the background and data point colors indicates correct classification, while discrepancies reveal misclassified resilience stages.
The examples demonstrate differences in the distinguishability of the Recovery stage across the considered systems.
In particular, the Recovery stage is clearly identifiable in the temporal evolution of system performance in Figures~\ref{fig:curveairport} and~\ref{fig:curvewifi}, whereas it is less pronounced and harder to distinguish from other resilience stages in Fig.~\ref{fig:curvelte}.

\label{sec:ltewifi}
Table \ref{tab:combined-results} presents the performance results of the investigated ML models for resilience stage recognition, evaluated separately on two datasets: the test set including IEEE 802.11 and LTE measurements and the Airport dataset. Model performance was assessed using class-specific Precision, Recall, and F1-score metrics for each resilience stage. The Airport dataset, originating from a different domain, was used to assess the models' generalization capabilities, particularly their ability to recognize resilience stages in a domain-agnostic and threat-agnostic manner.
\begin{figure*}[h] 
     \centering
     \subfloat[RF\label{fig:confmatrix-rf}]{
         \includegraphics[width=0.25\linewidth]{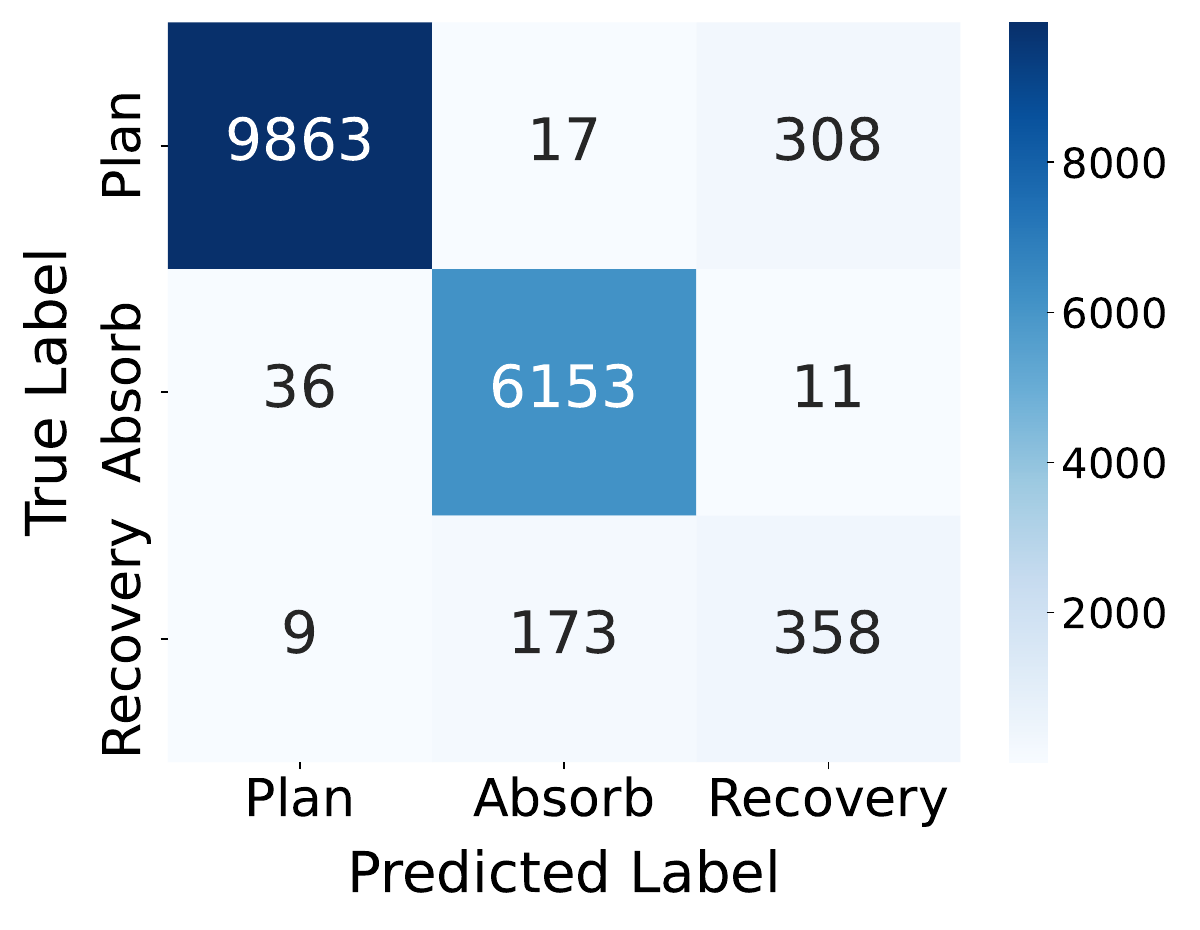}
     }
     \subfloat[SVM\label{fig:confmatrix-svm}]{
         \includegraphics[width=0.25\linewidth]{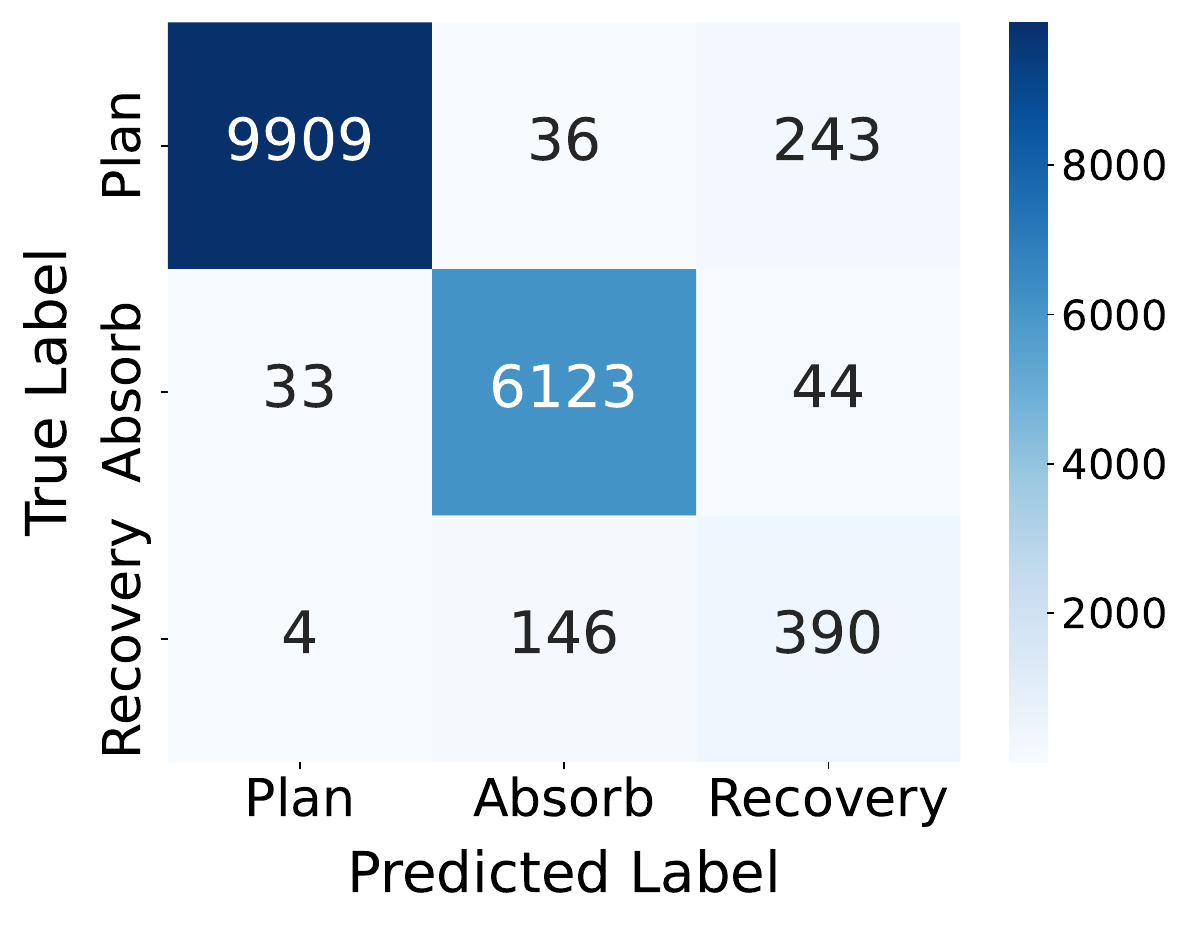}
     } 
     \subfloat[XGBoost\label{fig:confmatrix-xgb}]{
         \includegraphics[width=0.25\linewidth]{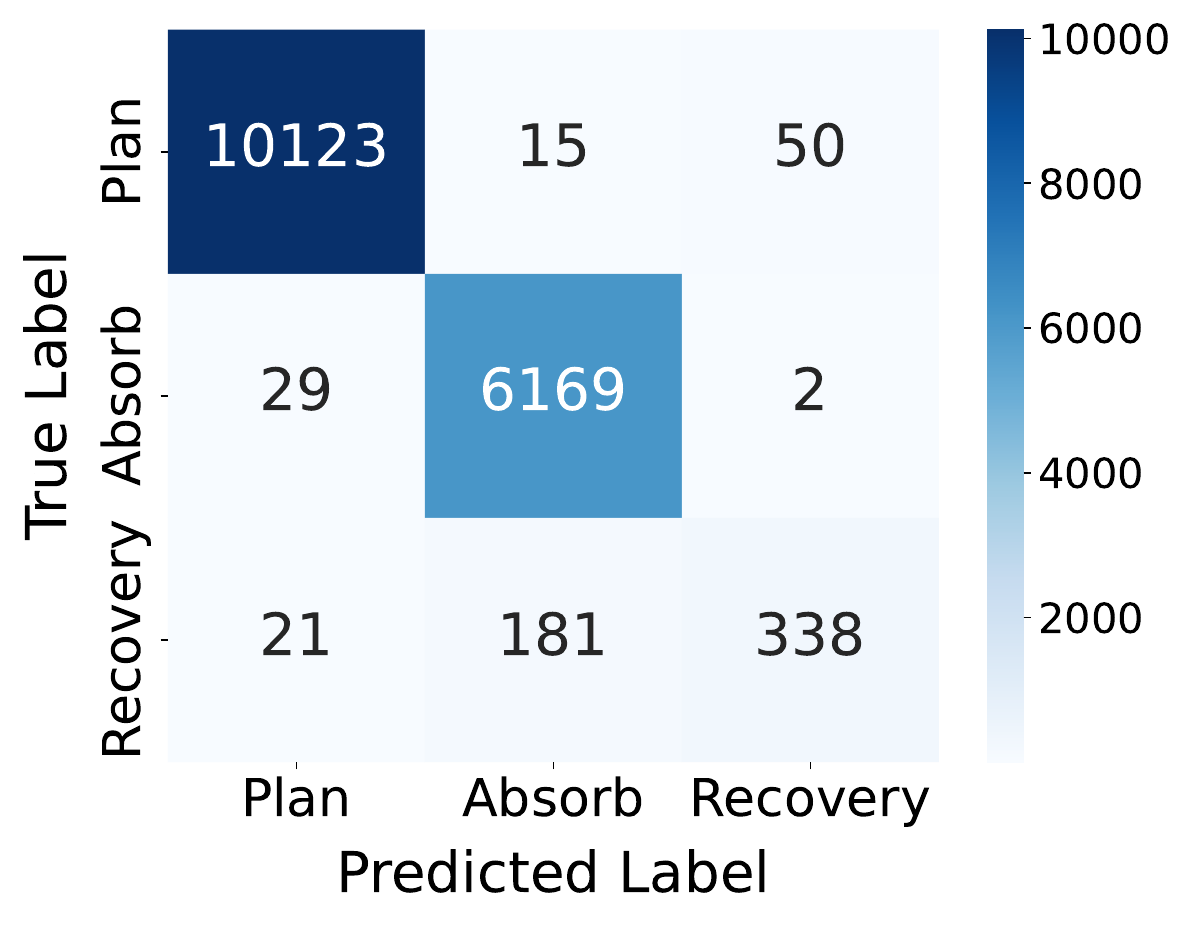}
     }
     \subfloat[LightGBM\label{fig:confmatrix-lgbm}]{
         \includegraphics[width=0.25\linewidth]{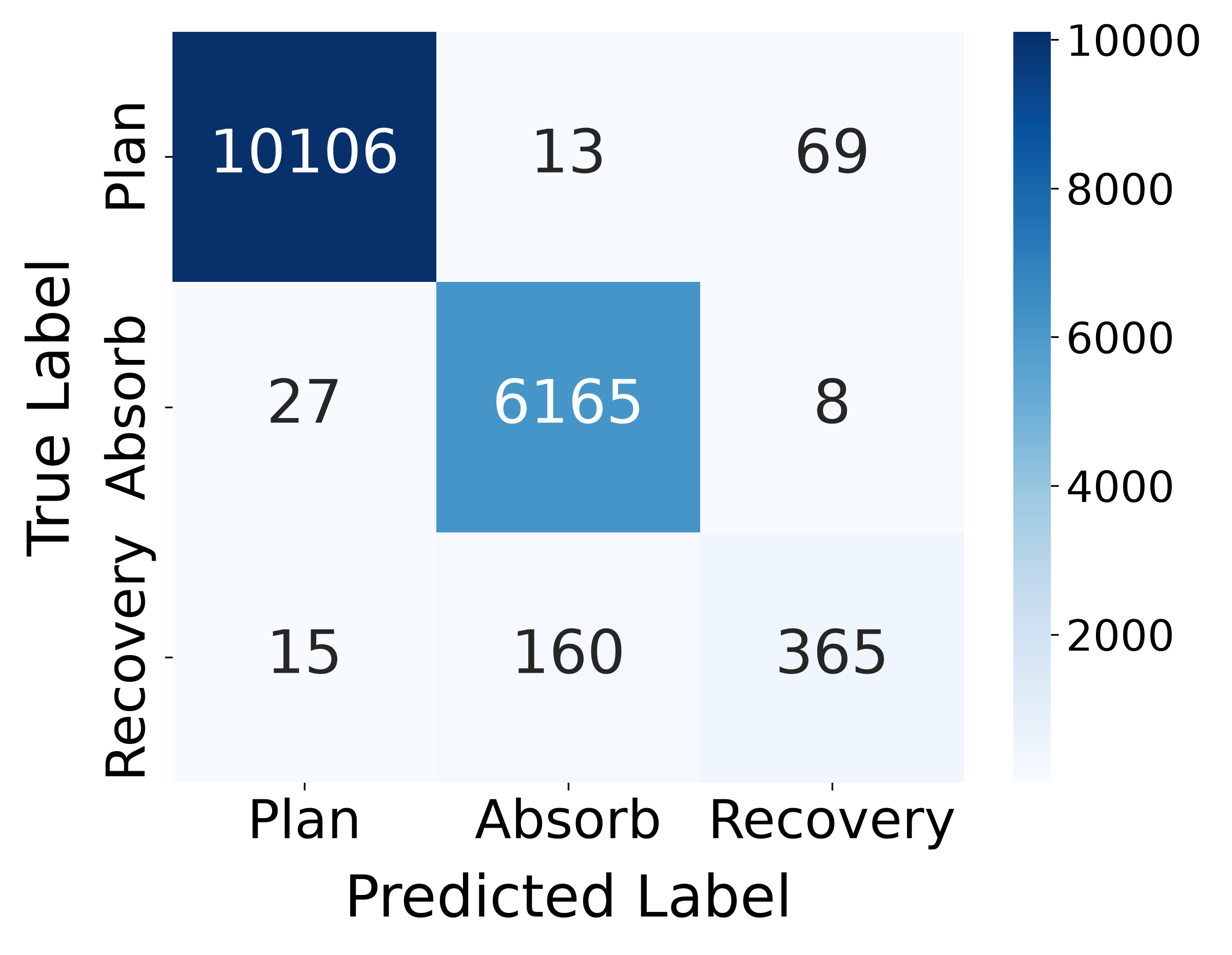}
     }   
     \caption{Confusion matrices for the RF, SVM, XGBoost and LightGBM models for IEEE 802.11 \& LTE datasets.}
     \label{fig:confusion-matrices}
\end{figure*}

\begin{figure*}[h] 
     \subfloat[RF\label{fig:confmatrix-rf-airport}]{
         \includegraphics[width=0.25\linewidth]{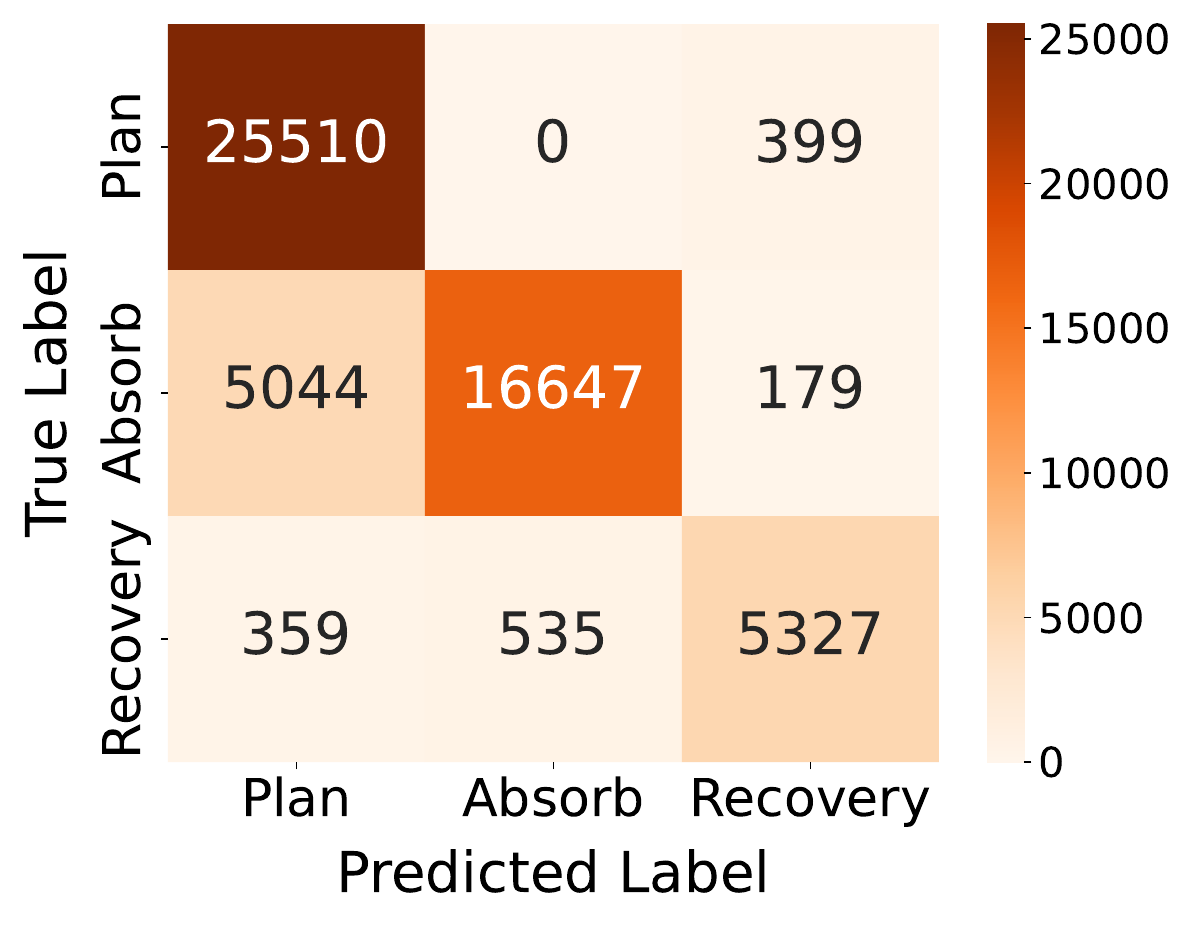}
     }
     \subfloat[SVM\label{fig:confmatrix-svm-airport}]{
         \includegraphics[width=0.25\linewidth]{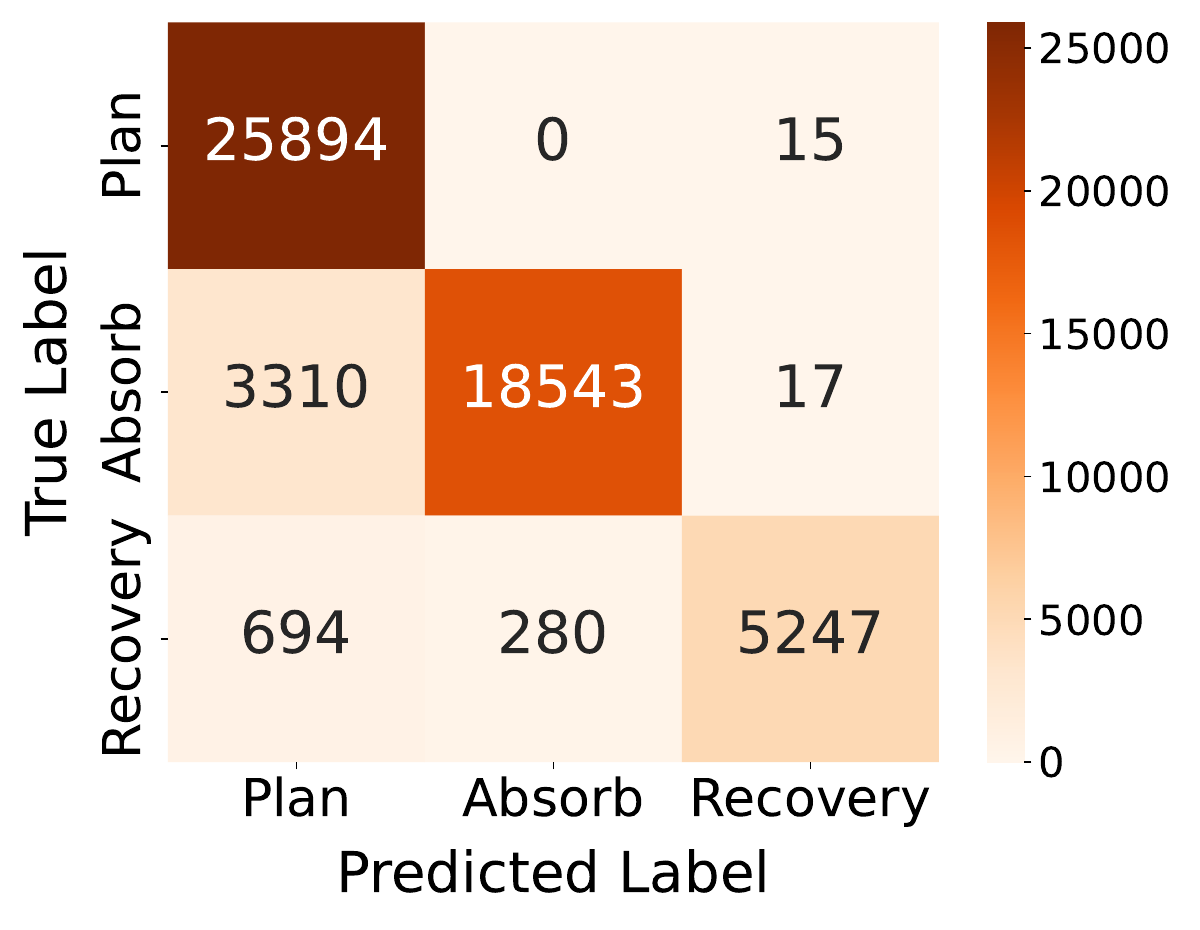}
     }
     \subfloat[XGBoost\label{fig:confmatrix-xgb-airport}]{
         \includegraphics[width=0.25\linewidth]{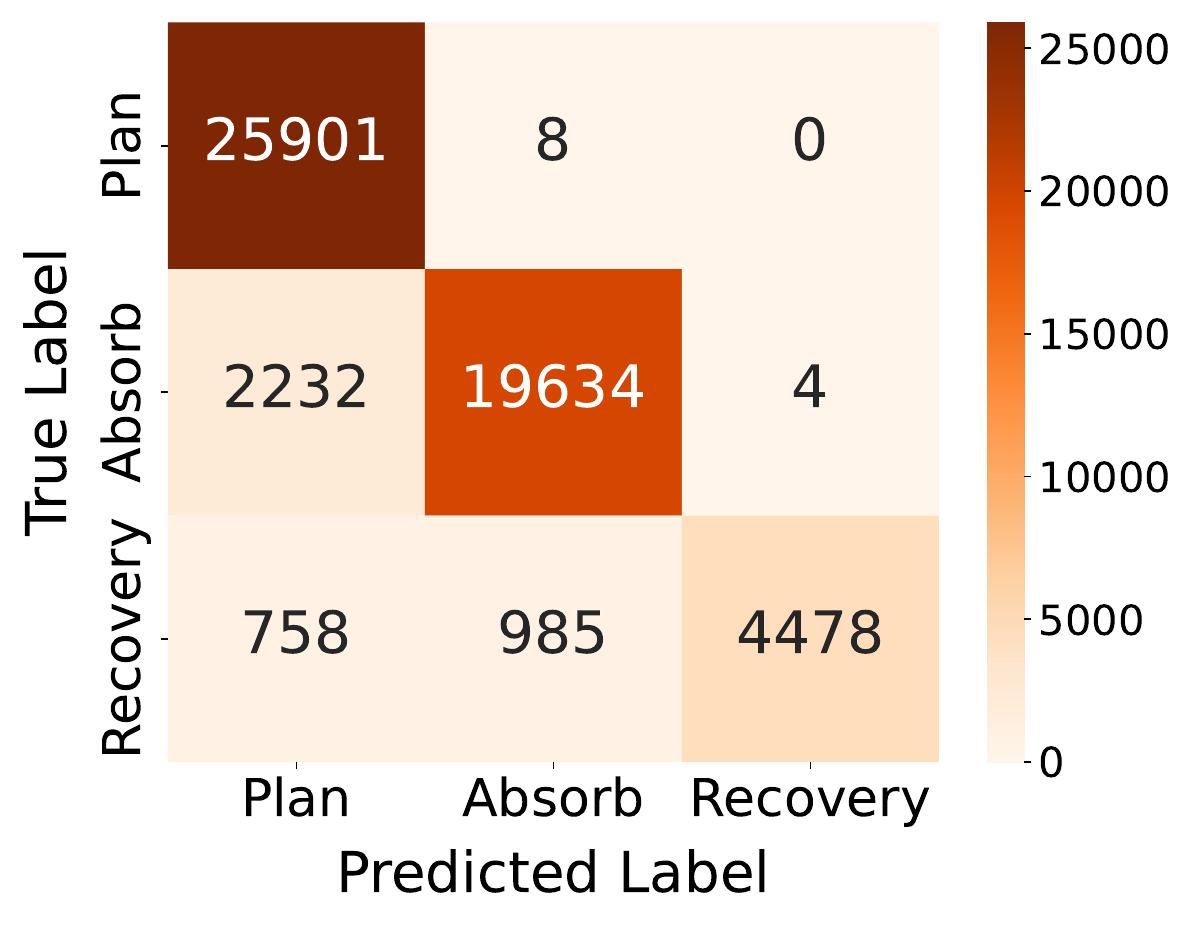}
     }
     \subfloat[LightGBM\label{fig:confmatrix-lgbm-airport}]{
         \includegraphics[width=0.25\linewidth]{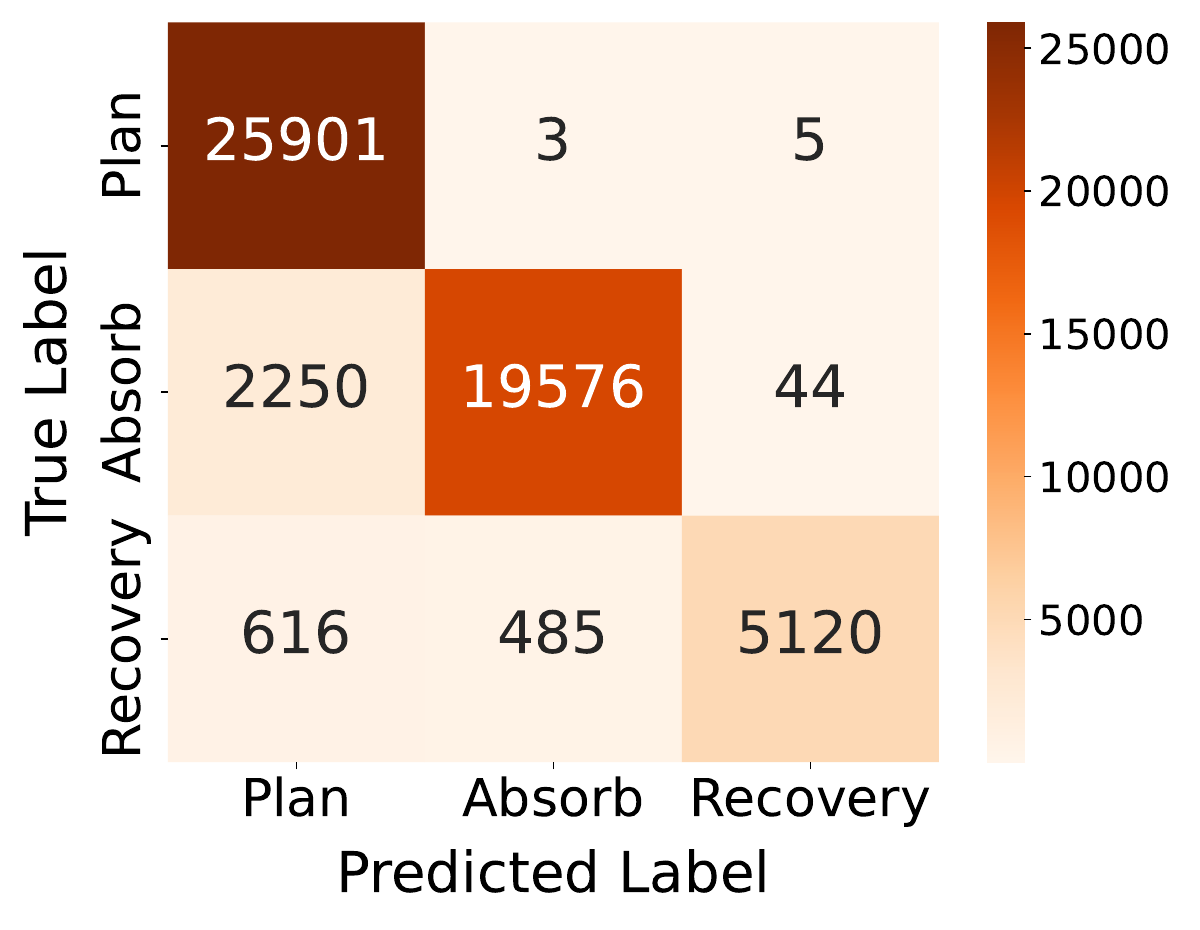}
     }   
     \caption{Confusion matrices for the RF, SVM, XGBoost and LightGBM models for the Airport dataset.}
     \label{fig:confusion-matrices-airport}
\end{figure*}

\begin{table*}[tbp]
\centering
\small
\caption{Performance of the ML models on the IEEE 802.11 \& LTE test set and on the Airport external validation dataset.}
\label{tab:combined-results}
\begin{tabular*}{\textwidth}{@{\extracolsep{\fill}} ll ccc ccc @{}}
\toprule
\multirow{2}{*}{\textbf{Model}} & \multirow{2}{*}{\textbf{Stage}} & \multicolumn{3}{c}{\textbf{IEEE 802.11 \& LTE (test set)}} & \multicolumn{3}{c}{\textbf{Airport (external validation set)}} \\
\cmidrule(lr){3-5} \cmidrule(lr){6-8}
& & \textbf{Precision (\%)} & \textbf{Recall (\%)} & \textbf{F1-Score (\%)} & \textbf{Precision (\%)} & \textbf{Recall (\%)} & \textbf{F1-Score (\%)} \\
\midrule

\multirow{3}{*}{\textbf{RF}}
& Plan    & 99.55 & 96.81 & 98.16 & 82.52 & 98.46 & 89.79 \\
& Absorb  & 97.00 & 99.24 & 98.11 & 96.89 & 76.12 & 85.26 \\
& Recover & 52.88 & 66.30 & 58.83 & 90.21 & 85.63 & 87.86 \\
\midrule

\multirow{3}{*}{\textbf{SVM}}
& Plan    & 99.63 & 97.26 & 98.43 & 86.61 & 99.94 & 92.80 \\
& Absorb  & 97.11 & 98.76 & 97.93 & 98.51 & 84.79 & 91.14 \\
& Recover & 57.61 & 72.22 & 64.09 & 99.39 & 84.34 & 91.25 \\
\midrule

\multirow{3}{*}{\textbf{XGBoost}}
& Plan    & 99.51 & 99.36 & 99.44 & 89.65 & 99.97 & 94.53 \\
& Absorb  & 96.92 & 99.50 & 98.19 & 95.19 & 89.78 & 92.40 \\
& Recover & 86.67 & 62.59 & 72.69 & 99.91 & 71.98 & 83.68 \\
\midrule

\multirow{3}{*}{\textbf{LightGBM}}
& Plan    & 99.59 & 99.20 & 99.39 & 90.04 & 99.97 & 94.74 \\
& Absorb  & 97.27 & 99.44 & 98.34 & 97.57 & 89.51 & 93.37 \\
& Recover & 82.58 & 67.59 & 74.34 & 99.05 & 82.30 & 89.90 \\
\bottomrule
\end{tabular*}
\end{table*}
\begin{table}[h]
\centering
\caption{Mean and standard deviation of performance metrics (in \%) from 5-fold cross-validation on the IEEE 802.11 \& LTE dataset.}
\label{tab:cross-val-results}
\begin{tabularx}{\linewidth}{l l X X X}
\hline
\textbf{Model} & \textbf{Stage} & \textbf{Precision (\%)} & \textbf{Recall (\%)} & \textbf{F1-Score (\%)}\\
\hline

\multirow{3}{*}{\textbf{RF}}
& Plan & 99.54 $\pm$ 0.22 & 97.09 $\pm$ 0.31 & 98.30 $\pm$ 0.16 \\
& Absorb    & 96.78 $\pm$ 0.21 & 99.38 $\pm$ 0.40 & 98.06 $\pm$ 0.16 \\
& Recover & 59.37 $\pm$ 4.62 & 67.91 $\pm$ 2.67 & 63.32 $\pm$ 3.62 \\
\hline

\multirow{3}{*}{\textbf{SVM}}
& Plan & 99.54 $\pm$ 0.25 & 97.04 $\pm$ 0.37 & 98.27 $\pm$ 0.12 \\
& Absorb    & 97.10 $\pm$ 0.26 & 98.56 $\pm$ 0.83 & 97.82 $\pm$ 0.39 \\
& Recover & 57.61 $\pm$ 3.96 & 73.43 $\pm$ 1.80 & 64.52 $\pm$ 3.01 \\
\hline

\multirow{3}{*}{\textbf{XGBoost}}
& Plan & 99.51 $\pm$ 0.18 & 99.28 $\pm$ 0.22 & 99.40 $\pm$ 0.11 \\
& Absorb    & 96.72 $\pm$ 0.22 & 99.60 $\pm$ 0.22 & 98.14 $\pm$ 0.10 \\
& Recover & 87.83 $\pm$ 3.54 & 64.18 $\pm$ 1.72 & 74.15 $\pm$ 2.28 \\
\hline

\multirow{3}{*}{\textbf{LightGBM}}
& Plan & 99.57 $\pm$ 0.18 & 99.13 $\pm$ 0.17 & 99.35 $\pm$ 0.06 \\
& Absorb    & 97.04 $\pm$ 0.17 & 99.47 $\pm$ 0.26 & 98.24 $\pm$ 0.09 \\
& Recover & 84.17 $\pm$ 3.05 & 68.65 $\pm$ 1.98 & 75.61 $\pm$ 2.30 \\
\hline
\end{tabularx}
\end{table}
On the test set, all evaluated models performed comparably in recognizing the Plan and Absorb stages, with F1-scores above 97\%, indicating that these two classes were consistently identified across models. The highest F1-score for the Plan stage was obtained by XGBoost ($>$99\%), while LightGBM achieved the highest F1-score for the Absorb stage ($>$98\%).

The evaluated ML models showed more pronounced differences in the Recover stage. The F1-score ranged from 58.8\% for RF to 74.3\% for LightGBM, while SVM and XGBoost achieved 64\% and 72.7\%, respectively. Although XGBoost obtained the highest Precision for this class (86.7\%), LightGBM achieved the highest Recall (67.6\%) and the highest F1-score, indicating the best overall balance between Precision and Recall for Recover-stage recognition. Overall, the Recover stage proved to be the most challenging to classify, with consistently lower performance than the Plan and Absorb stages across all evaluated ML models.

The lower classification performance observed for the Recover stage is consistent with the confusion matrices shown in Figure \ref{fig:confusion-matrices}. Across all evaluated ML models, samples from the Recover stage were more often misclassified as Absorb than as Plan, indicating that the main source of classification error stems from limited separability between the Absorb and Recover stages. Specifically, the Recover stage was misclassified as Absorb about 9--37 times more often than as Plan. This suggests a greater overlap in the feature distributions of the Absorb and Recover stages, resulting in a less distinct boundary between these classes. Figure \ref{fig:curvelte} illustrates this effect.

The IEEE 802.11 \ref{fig:curvewifi} and Airport \ref{fig:curveairport} examples show a clearly distinguishable Recovery stage, whereas Fig.~\ref{fig:curvelte} shows a much less distinct Recovery stage that largely blends with the Absorb stage. Recovery samples correspond to observations occurring after the disruption, i.e., after the Absorb stage, and do not exhibit the characteristics associated with the Adapt or Plan stages. Although their signal characteristics may remain similar to those observed during Absorb, their temporal position after the disruptive event distinguishes them as part of the Recovery stage.
Therefore, measurements collected during Absorb and Recovery stages may exhibit similar feature values, reducing their separability in the feature space. In contrast, the Plan stage is characterized by a distinct operating state, which likely contributes to its consistently high classification performance across all evaluated ML models.

Compared with the IEEE 802.11 \& LTE test set results, the Airport validation set (Table \ref{tab:combined-results}) showed noticeable shifts in classification performance, although all models still generalized reasonably well to the new domain.
For the Plan and Absorb stages, overall performance slightly decreased. While the models maintained exceptionally high Recall for the Plan stage (up to 99.97\% for XGBoost and LightGBM), this came at the cost of lower Precision, which dropped to between 82.5\% (RF) and 90\% (LightGBM). Conversely, for the Absorb stage, the models maintained high Precision (above 95\% for all models except RF) but experienced a notable drop in Recall (ranging from 76.1\% to 89.8\%). This inverse relationship suggests that, on the validation set, the models tended to over-predict the Plan stage, increasing false positives, likely by misclassifying Absorb instances. 

Interestingly, classification of the Recover stage, which proved most challenging in the test set (IEEE 802.11 \& LTE), improved significantly in the external validation set (Airport). The higher F1-score for the Recover stage in the Airport dataset is attributed to its longer and more gradual recovery process (see Figures \ref{fig:curveairport}, \ref{fig:airport-data}), while the lower recall for the Absorb stage may result from the gradual onset of the disruption, rather than differences in model performance. All evaluated models achieved F1-scores above 83\%, a substantial increase from the 58-74\% range observed in the test set. Most notably, Precision for the Recover stage surged across all models, exceeding 99\% for SVM, XGBoost, and LightGBM. This indicates that while the models were highly conservative in predicting the Recover stage on the unseen data, the predictions they did make were almost entirely correct. 

In terms of overall robustness on the validation set, LightGBM and SVM emerged as the most consistent models. LightGBM achieved the highest F1-scores for the Plan (94.7\%) and Absorb (93.4\%) stages, maintaining a strong balance between Precision and Recall. Meanwhile, SVM achieved the highest F1-score for the Recover stage (91.3\%), slightly outperforming LightGBM (89.9\%) and significantly outperforming XGBoost (83.7\%), which struggled with a lower Recall (72\%) for this specific class. 

Overall, the proposed ML approach demonstrated robust generalization when evaluated cross-domain on new, unseen data. Across the out-of-domain validation set, the evaluated models maintained strong predictive performance, yielding F1-scores ranging from 83.68\% (XGBoost for the Recover stage) up to 94.74\% (LightGBM for the Plan stage). A comparative analysis reveals a compelling shift in class separability between the datasets. In the baseline test set, the models easily distinguished the Plan and Absorb stages (F1-scores $>$ 97\%) but performed worse on the Recover stage (F1-scores between 58.83\% and 74.34\%). Conversely, on the unseen validation domain, the recognition of the previously challenging Recover stage improved significantly, reaching F1-scores up to 91.25\% (SVM) while maintaining robust overall performance. Overall, the validation results show that this dataset's distinct features enabled highly precise classification, with tree-based boosting algorithms and SVM showing strong generalization potential.

Table \ref{tab:cross-val-results} presents the mean and standard deviation of ML models' performance metrics: precision, recall, and F1-score (in \%) from 5-fold cross-validation on the whole IEEE 802.11 and LTE dataset. The dataset was divided into five approximately equal-sized subsets (folds). In each iteration, we used four folds (80\% of the data) for training, while the remaining fold (20\%) was used for validation. We repeated this procedure five times, with each fold serving as the validation set exactly once. We assessed final performance by averaging the evaluation metrics across all five iterations. The results reported in Table \ref{tab:cross-val-results} are supplementary. Their primary purpose is to assess the stability and consistency of the proposed ML models across the considered experiments. They therefore provide additional evidence on model robustness, complementing the results obtained from the final training procedure. The outcomes of the final model training, which constitute the main basis for the subsequent analysis, are reported separately in Table \ref{tab:combined-results}.

Overall, the cross-validation results presented in Table \ref{tab:cross-val-results} confirm that the performance trends observed on the test dataset are consistent across different data partitions. The low standard deviations obtained for the Plan and Absorb stages indicate stable classification performance. In contrast, the higher variability in the Recover stage is consistent with the greater classification difficulty reflected in the confusion matrices and the lower Recall values for this class.
For the Plan and Absorb stages, all evaluated models achieved consistently high performance with very low standard deviations. The F1-score standard deviation remained below 0.4 percentage points for all models, indicating that the classification performance was largely insensitive to the specific data partition. In contrast, larger standard deviations were observed for the Recover stage, with F1-score variability ranging from 2.28 p.p. (XGBoost) to 3.62 p.p RF). Despite this increased variability, the model ranking remained unchanged, with LightGBM achieving the highest mean F1-score (75.61\%$\pm$2.3), followed by XGBoost (74.15\%$\pm$2.28), SVM (64.52$\%\pm$3.01), and RF (63.32$\%\pm$3.62).

\subsection{Resilience Quantification}
Table \ref{tab:resilience-metrics-errors} presents the performance of the proposed approach, in which ML models are used to recognize resilience stages, and their predictions are subsequently used to calculate resilience metrics ($R_t$, RL, $\epsilon$, and $\Lambda$) according to their mathematical definitions. The accuracy of the calculated metrics was evaluated using Mean Absolute Error (MAE), Root Mean Squared Error (RMSE), and Mean Absolute Percentage Error (MAPE). MAPE is simply the average of the APE (Absolute Percentage Error) values calculated for each observation: 
$
\mathrm{MAPE} = \frac{100\%}{N}
\sum_{i=1}^{N}
\left|
\frac{y_{\mathrm{true},i} - y_{\mathrm{pred},i}}
{y_{\mathrm{true},i}}
\right|,$
where $y_{(pred,i)}$ is a predicted value, $y_{true,i}$ is a ground-truth, and $N$ is total amount of observations.
These measures capture complementary aspects of estimation error: MAE quantifies the average absolute deviation from reference values, RMSE assigns greater weight to larger deviations, and MAPE expresses errors relative to the reference values as a percentage. 

For the IEEE 802.11 and LTE test sets, all models achieved relatively low errors across the investigated resilience metrics. XGBoost and LightGBM generally outperformed RF and SVM, particularly in MAE and MAPE. XGBoost achieved the lowest MAPE for $R_t$ (0.82\%) and RL (0.35\%), whereas LightGBM provided the most accurate predictions for $\epsilon$ (3.08\%) and $\Lambda$ (0.9\%). Both LightGBM and XGBoost showed comparable MAE and RMSE, indicating their effectiveness in estimating resilience metrics under conditions similar to those in the training data.

Although $R_t$ estimation achieves exceptionally high accuracy on the IEEE 802.11 and LTE dataset ($\text{MAPE} < 1.6\%$ across all models, reaching $0.82\%$ for XGBoost and $0.87\%$ for LightGBM), prediction errors increase substantially on the Airport dataset, ranging from $9.56\%$ to $10.07\%$ for boosting-based models and from $13.38\%$ to $15\%$ for SVM and RF. This discrepancy can be attributed to differences in the temporal characteristics of system performance degradation. In the Airport dataset, transitions between stages, particularly in the failure stage, occur gradually (see Fig.~\ref{fig:curveairport}) rather than abruptly as in the IEEE 802.11 and LTE datasets (see Fig.~\ref{fig:curvelte}, \ref{fig:curvewifi}). Consequently, the model detects the onset of the Absorb stage with a slight delay.
As a result, the early steps of the Absorb stage are temporarily misclassified as Plan (see Fig. \ref{fig:curveairport}). This delay directly explains why Recall for the Absorb stage on the Airport set reaches at best about $89\%$, about $10$ percentage points lower than the $\approx 99\%$ achieved on the IEEE 802.11 and LTE dataset.

Consequently, the model misses the early edge of the Absorb stage. Because this edge is where system performance remains highest, the area under the curve is also largest there. Omitting this substantial boundary area significantly depresses the calculated mean retained functionality, which directly causes the increased MAPE observed for $R_t$ on the Airport dataset.

A similar but much smaller increase in error is observed for $RL$, also indicating worse performance on the Airport dataset. The models achieve slightly higher MAPE on the Airport dataset ($1.65\%$ for XGBoost and $1.67\%$ for LightGBM) than on the IEEE 802.11 and LTE datasets ($0.35\%$–$0.37\%$). However, this increase is much smaller than for $R_t$.
Although $R_t$ and $RL$ are related, they measure different aspects of the disturbance. The $RL$ metric measures the total performance loss over time. At the beginning of the Absorb stage, congestion is still low, so the performance loss is small (see Figure \ref{fig:rt-metric}). If the model detects the start of this stage slightly late, it misses only a small part of the total loss. Most of the loss occurs later in the disruption, and the model captures it accurately. Therefore, small errors in detecting the start of the Absorb stage have little effect on $RL$. This explains why the MAPE for $RL$ remains below $1.7\%$ on the Airport dataset.

\begin{table*}[htpb]
\centering
\small 
\caption{Comparison of MAE, RMSE, and MAPE metrics between the IEEE 802.11 \& LTE dataset test set and  Airport external validation set.}
\label{tab:resilience-metrics-errors}
\begin{tabular*}{\textwidth}{@{\extracolsep{\fill}} ll ccc ccc @{}}
\toprule
\multirow{2}{*}{\textbf{Resilience Metric}} & \multirow{2}{*}{\textbf{Model}} & \multicolumn{3}{c}{\textbf{IEEE 802.11 \& LTE (test set)}} & \multicolumn{3}{c}{\textbf{Airport (external validation set)}} \\
\cmidrule(lr){3-5} \cmidrule(lr){6-8}
& & \textbf{MAE} & \textbf{RMSE} & \textbf{MAPE (\%)} & \textbf{MAE} & \textbf{RMSE} & \textbf{MAPE (\%)} \\
\midrule

\multirow{4}{*}{$R_t$ \cite{de2025quantitative,LIU2021107868}}
& RF       & 0.011 & 0.015 & 1.54 & 0.07 & 0.08 & 15.00 \\
& SVM      & 0.010 & 0.015 & 1.51 & 0.07 & 0.07 & 13.38 \\
& XGBoost  & 0.005 & 0.013 & 0.82 & 0.05 & 0.05 & 10.07 \\
& LightGBM & 0.005 & 0.013 & 0.87 & 0.05 & 0.05 & 9.56 \\
\midrule

\multirow{4}{*}{RL \cite{bruneau2003framework,panteli2017}}
& RF       & 0.05 & 0.11 & 0.55 & 2.94 & 3.36 & 5.55 \\
& SVM      & 0.04 & 0.08 & 0.50 & 1.60 & 2.00 & 3.08 \\
& XGBoost  & 0.03 & 0.09 & 0.35 & 0.89 & 1.21 & 1.65 \\
& LightGBM & 0.03 & 0.08 & 0.37 & 0.90 & 1.23 & 1.67 \\
\midrule

\multirow{4}{*}{$\epsilon$  \cite{app13116452,panteli2017}}
& RF       & 1.15 & 1.43 & 3.78 & 17.73 & 19.19 & 21.65 \\
& SVM      & 1.27 & 1.48 & 4.12 & 11.44 & 13.17 & 13.92 \\
& XGBoost  & 1.09 & 1.18 & 3.51 & 5.06 & 6.67 & 6.07 \\
& LightGBM & 0.97 & 1.09 & 3.08 & 6.78 & 8.16 & 8.24 \\
\midrule

\multirow{4}{*}{$\Lambda$ \cite{app13116452,panteli2017}}
& RF       & 0.004 & 0.010 & 1.11 & 0.11 & 0.12 & 22.10 \\
& SVM      & 0.004 & 0.012 & 1.35 & 0.08 & 0.09 & 15.01 \\
& XGBoost  & 0.003 & 0.011 & 0.99 & 0.06 & 0.06 & 11.37 \\
& LightGBM & 0.003 & 0.011 & 0.90 & 0.06 & 0.06 & 10.60 \\
\bottomrule
\end{tabular*}
\end{table*}

\begin{figure*}[htpb] 
    \centering
    \subfloat[$R_t$ \label{fig:rt}]{
        \includegraphics[width=0.25\linewidth]{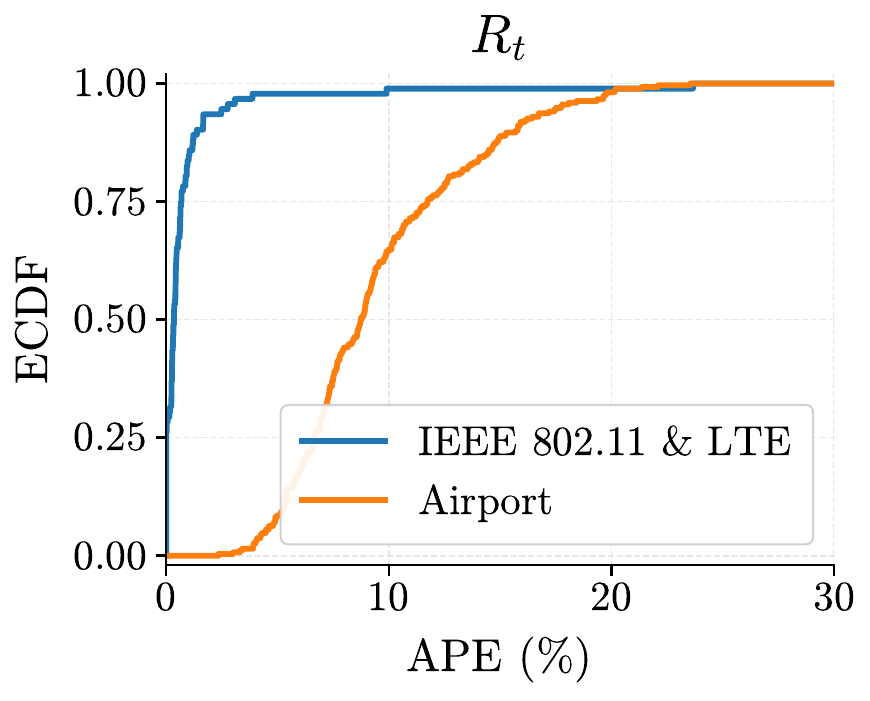}
    } 
    \subfloat[RL\label{fig:RL}]{
        \includegraphics[width=0.25\linewidth]{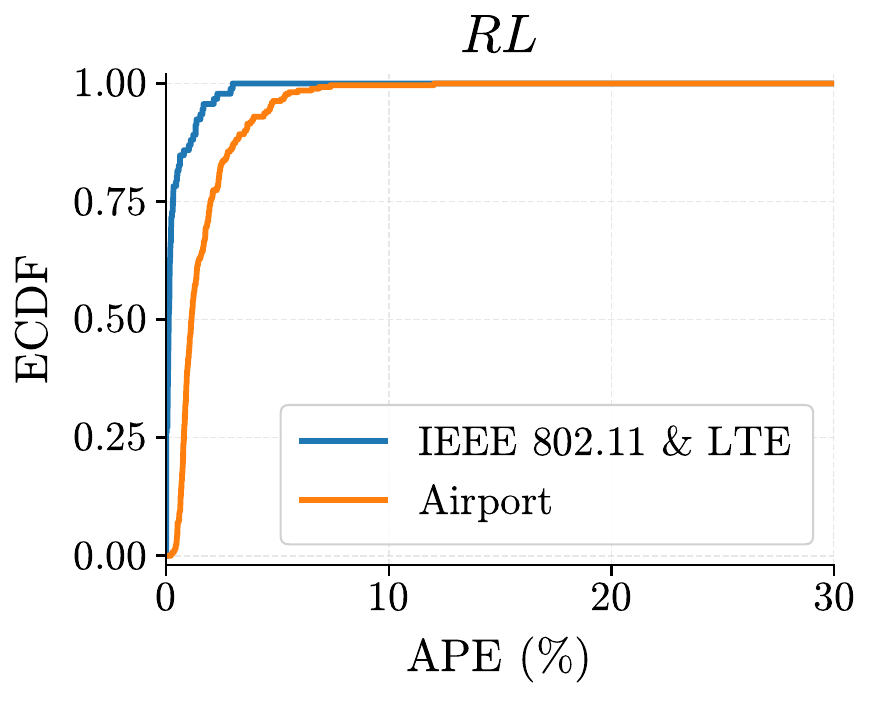}
    } 
    \subfloat[$\epsilon$\label{fig:eps}]{
        \includegraphics[width=0.25\linewidth]{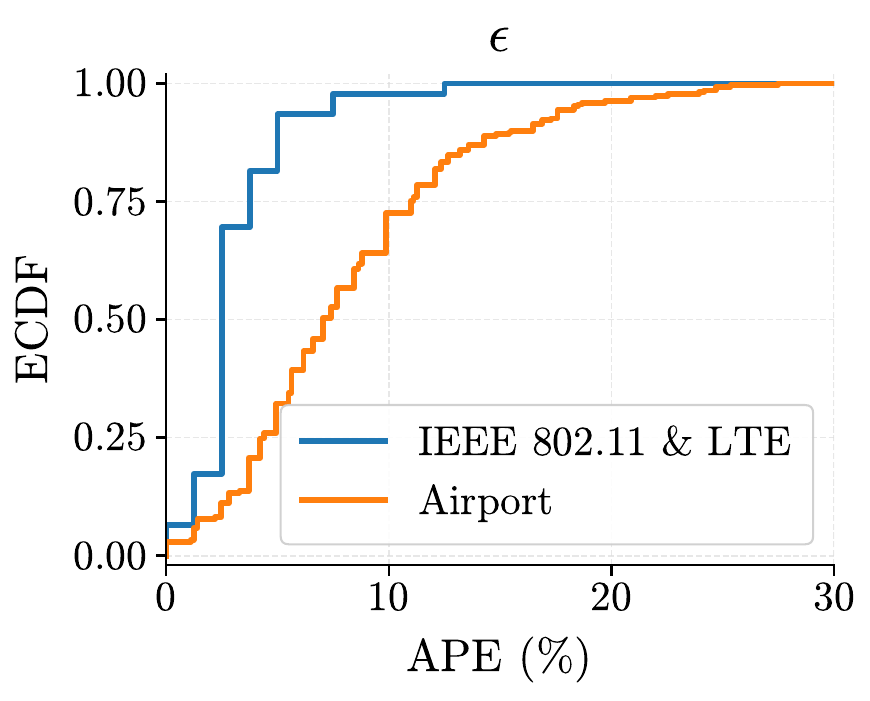}
    } 
    \subfloat[$\Lambda$\label{fig:lambda}]{
        \includegraphics[width=0.25\linewidth]{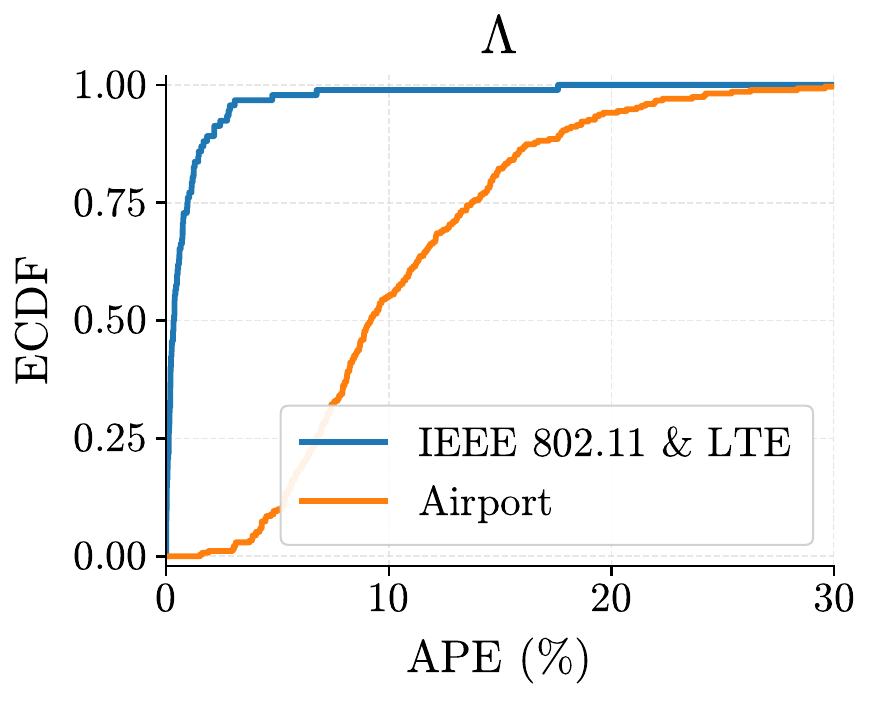}
    }
    \caption{Empirical Cumulative Distribution Functions (ECDFs) of the Absolute Percentage Error (APE) for the resilience metrics $R_t$, RL, $\epsilon$, and $\Lambda$, comparing LightGBM model-based results on the IEEE 802.11 \& LTE test set and Airport external validation set.}
    \label{fig:ecdf-ape-metrics}
\end{figure*}

Figure \ref{fig:ecdf-ape-metrics} presents the empirical cumulative distribution functions (ECDFs) of the Absolute Percentage Error (APE) for each resilience metric ($R_t$, $RL$, $\epsilon$, and $\Lambda$), obtained using the LightGBM model as a representative example. For each metric, the APE distributions are shown separately for the IEEE 802.11 \& LTE and airport datasets, allowing a comparison of prediction accuracy across the two datasets. The curves illustrate the proportion of predictions with an APE below a given threshold, providing insight into the distribution of prediction errors for each resilience metric. Table \ref{tab:ecdf-ape-lightgbm} includes the detailed numerical values.

Table \ref{tab:ecdf-ape-lightgbm} presents the ECDF of APE for resilience metrics calculated with LightGBM, one of the best-performing models. On the IEEE 802.11 and LTE test set, at least 97.8\% of estimates exhibited errors below 10\% across all resilience metrics. On the unseen Airport validation set, RL demonstrated the strongest generalization, with 99.6\% of estimates below 10\% APE, followed by $\epsilon$ (72.6\%), $R_t$ (64.8\%), and $\Lambda$ (55.2\%). These results highlight LightGBM's potential for cross-domain resilience assessment, despite reduced accuracy under domain shifts.

\begin{table*}[hbtp]
\caption{Empirical cumulative distribution function (ECDF) of APE (\%) for resilience metrics under the best-performing ML model (LightGBM).}
\label{tab:ecdf-ape-lightgbm}
\centering
\small
\begin{tabular*}{\textwidth}{@{\extracolsep{\fill}} llcccccc @{}}
\toprule
Resilience Metric & Dataset & $< 3\%$ & $< 5\%$ & $< 10\%$ & $< 15\%$ & $< 20\%$ & $< 25\%$ \\
\midrule
\multirow{2}{*}{$R_t$ \cite{de2025quantitative,LIU2021107868}} 
 & IEEE 802.11 \& LTE & 95.7 & 97.8 & 98.9 & 98.9 & 98.9 & 100 \\
 & Airport            & 0.37 & 8.5 & 64.8 & 88.5 & 98.2 & 100 \\
\midrule
\multirow{2}{*}{$RL$ \cite{bruneau2003framework,panteli2017}} 
 & IEEE 802.11 \& LTE & 100 & 100 & 100 & 100 & 100 & 100 \\
 & Airport            & 87.0 & 96.3 & 99.6 & 100 & 100 & 100 \\
\midrule
\multirow{2}{*}{$\epsilon$ \cite{app13116452,panteli2017}} 
 & IEEE 802.11 \& LTE & 69.6 & 81.5 & 97.8 & 100 & 100 & 100 \\
 & Airport            & 13.3 & 32.2 & 72.6 & 89.3 & 96.3 & 99.3 \\
\midrule
\multirow{2}{*}{$\Lambda$ \cite{app13116452,panteli2017}} 
 & IEEE 802.11 \& LTE & 95.7 & 97.8 & 98.9 & 98.9 & 100 & 100 \\
 & Airport            & 1.5 & 9.6 & 55.2 & 82.2 & 94.1 & 98.2 \\
\bottomrule
\end{tabular*}
\end{table*}

\begin{figure*}[h] 
    \centering
    \subfloat[$R_t$\label{fig:scatter-rt}]{
        \includegraphics[width=0.25\linewidth]{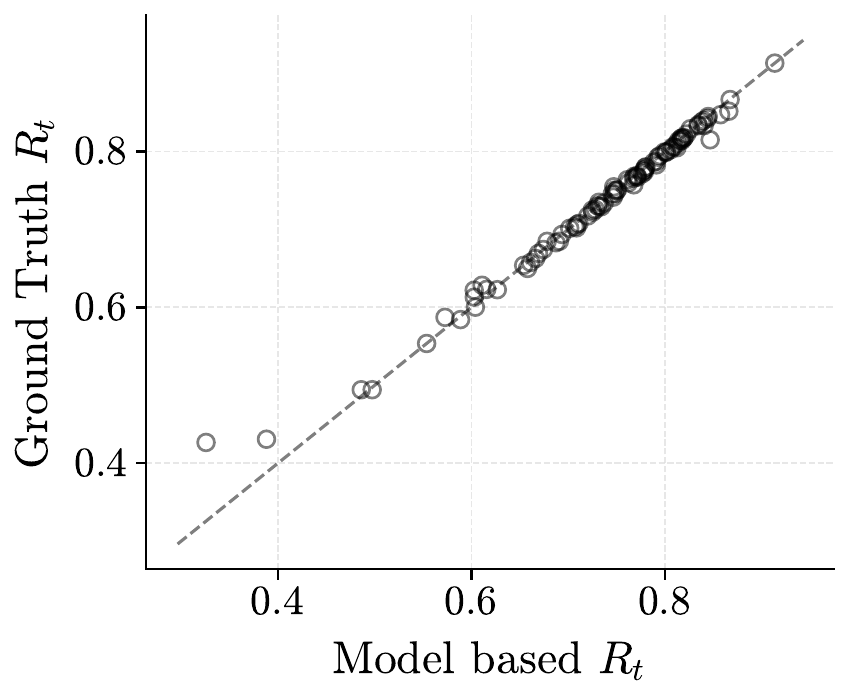}
    } 
    \subfloat[RL\label{fig:scatter-RL}]{
        \includegraphics[width=0.25\linewidth]{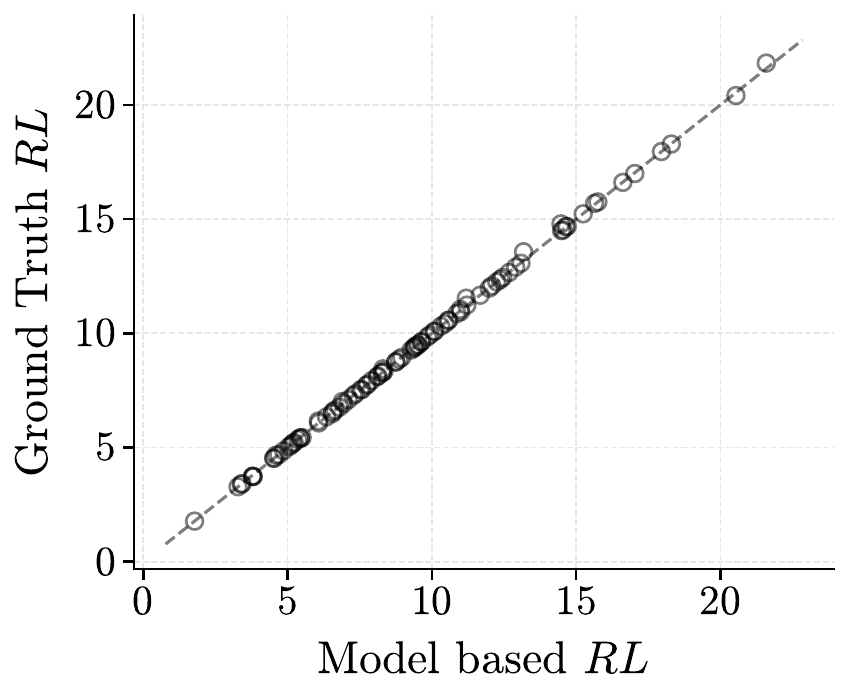}
    } 
    \subfloat[$\epsilon$\label{fig:scatter-eps}]{
        \includegraphics[width=0.25\linewidth]{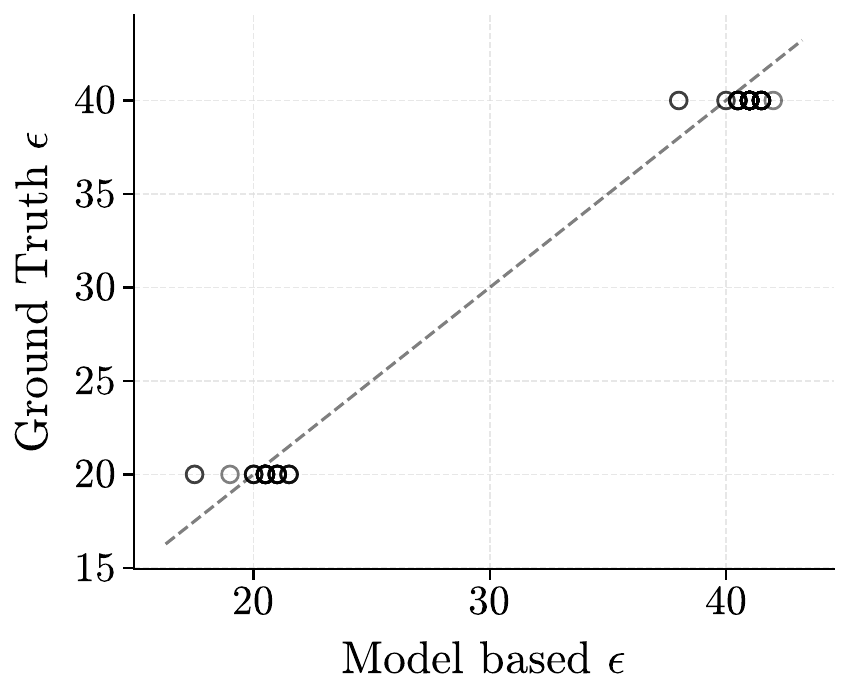}
    } 
    \subfloat[$\Lambda$\label{fig:scatter-lambda}]{
        \includegraphics[width=0.25\linewidth]{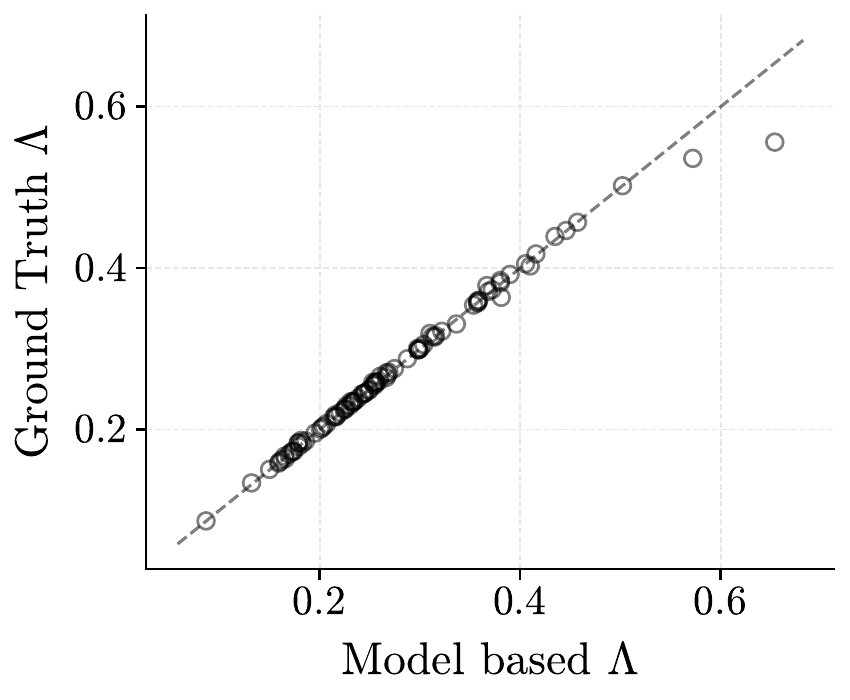}
    }
    \caption{Scatter plots of model-based vs ground truth for all metrics $R_t$, RL, $\epsilon$, $\Lambda$ for IEEE 802.11 \& LTE dataset.}
    \label{fig:scatter-testset}
\end{figure*}

\begin{figure*}[h] 
    \centering
    \subfloat[$R_t$\label{fig:scatter-rt-air}]{
        \includegraphics[width=0.25\linewidth]{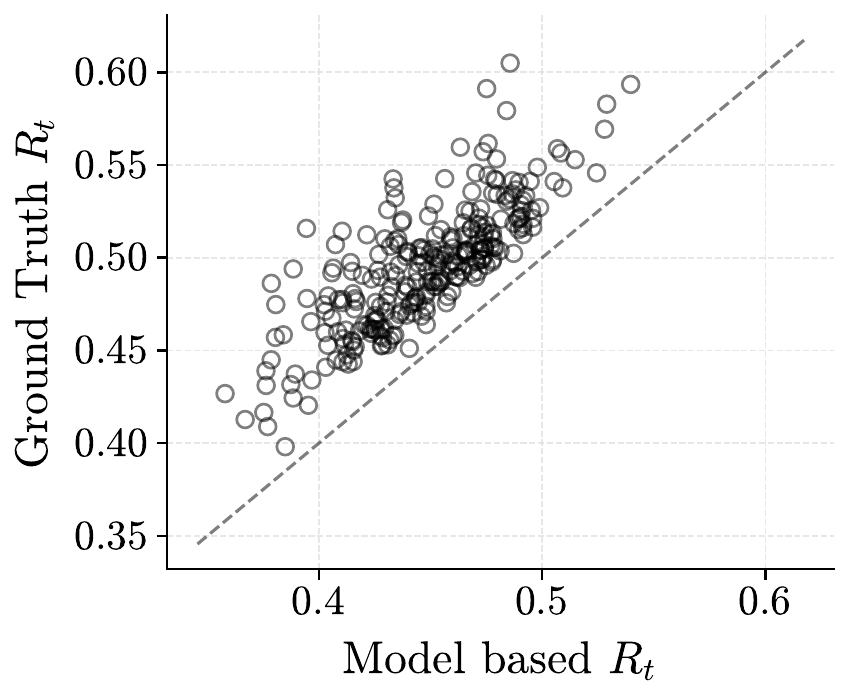}
    } 
    \subfloat[RL\label{fig:scatter-RL-air}]{
        \includegraphics[width=0.25\linewidth]{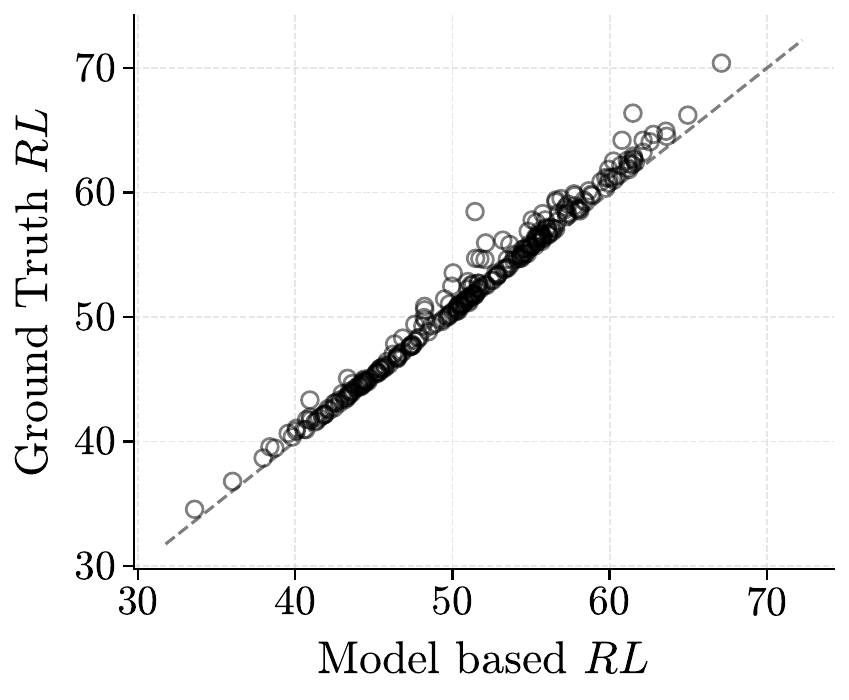}
    } 
    \subfloat[$\epsilon$\label{fig:scatter-eps-air}]{
        \includegraphics[width=0.25\linewidth]{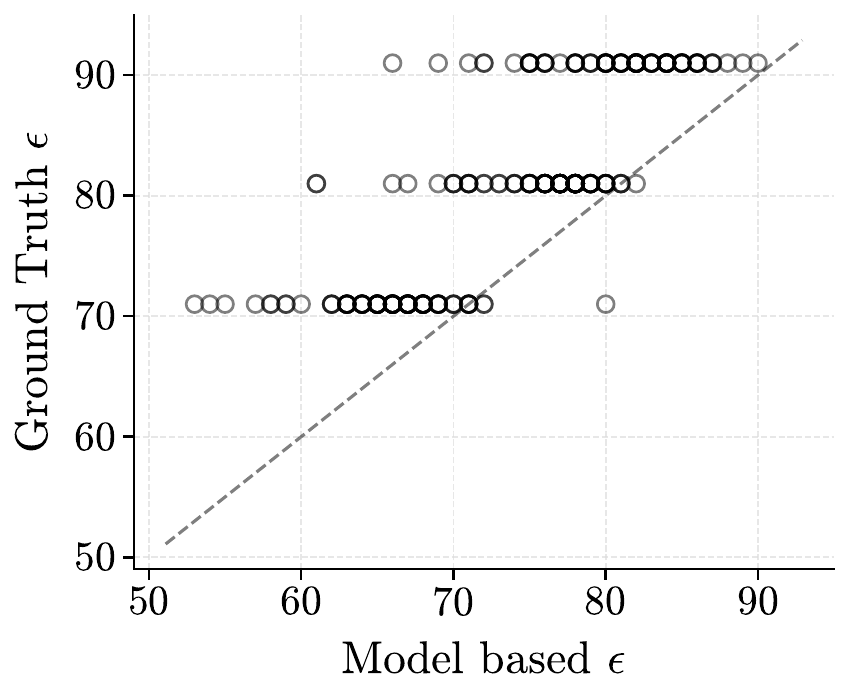}
    } 
    \subfloat[$\Lambda$\label{fig:scatter-lambda-air}]{
        \includegraphics[width=0.25\linewidth]{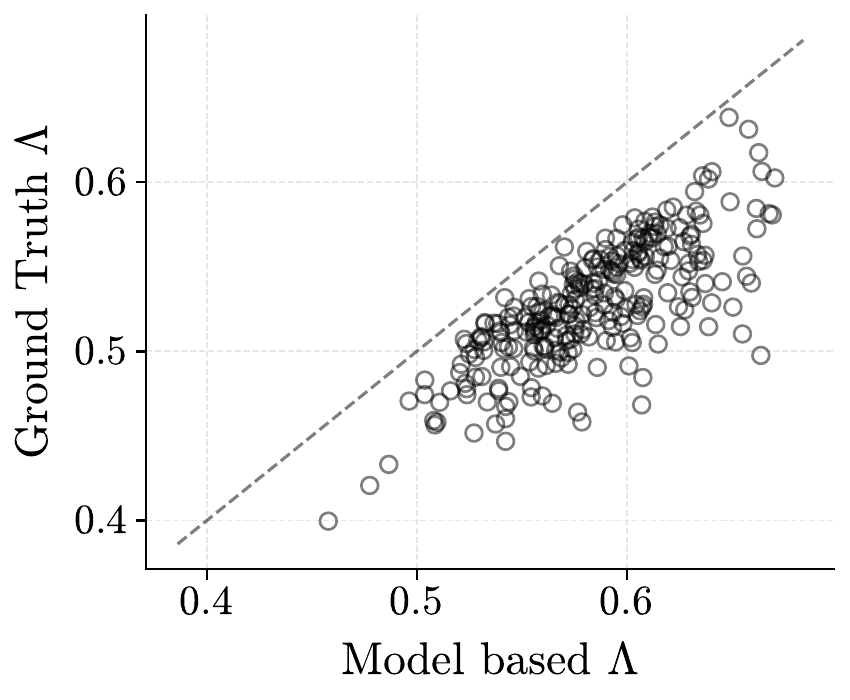}
    }
    \caption{Scatter plots of model-based vs. ground truth for all metrics $R_t$, RL, $\epsilon$, $\Lambda$ for the Airport dataset.}
    \label{fig:scatter-ariport}
\end{figure*}

Figures \ref{fig:scatter-testset} and \ref{fig:scatter-ariport} compare ground-truth resilience metrics ($R_t$, $RL$, $\epsilon$, and $\Lambda$) with those calculated from ML model outputs. LightGBM was selected as a representative model to illustrate predictive performance. The diagonal reference line on the charts represents perfect agreement between the ground truth and values calculated from the ML model's classification outcome.
Points close to this line indicate small estimation errors, whereas larger deviations correspond to larger errors.


The scatter plots for the IEEE 802.11 \& LTE dataset show strong predictive accuracy, with most predictions closely following the diagonal reference line, indicating close agreement between the ground-truth and ML model-based values. This performance is expected, as the test data originate from the same domain as the data used to train the model. In contrast, predictions for the Airport dataset show larger deviations from the ground-truth values, reflecting the challenges of applying the model to data from a different domain. Nevertheless, the results show promising predictive performance for $RL$, a key resilience metric, as it reflects the overall impact of performance degradation across the Absorb and Recover stages. For the remaining resilience metrics, we observe systematic biases: $R_t$ tends to be underestimated, whereas $\Lambda$ is generally overestimated. This is directly attributable to the delayed identification of the Absorb stage in the time series, with its initial portion being misclassified as the Plan stage (see Figures \ref{fig:curveairport}, \ref{fig:confusion-matrices-airport}). 
Despite these biases, the predicted values remain approximately aligned with the diagonal reference line, suggesting that the model captures the overall trends and relationships in the data, even when applied to a different domain.

For the $\epsilon$ resilience metric (Fig.~\ref{fig:scatter-eps}, \ref{fig:scatter-eps-air}), the scatter plot exhibits a distinct pattern, with data points concentrated along a few horizontal bands. This distribution results from the limited number of unique ground-truth values, which causes multiple observations to share the same reference value despite differences in their model-based predictions. Consequently, the observed clustering reflects the discrete nature of the ground-truth data rather than an artifact of the prediction model. The points' proximity to the diagonal reference line indicates agreement between predicted and actual values.

\section{Conclusions}
\label{sec:conclusions}
This paper introduced a domain- and threat-agnostic ML-based approach to automatically recognize PARA (Plan/Prepare, Absorb, Recover, Adapt) resilience stages in system-performance time series for further quantitative resilience analysis across diverse systems.
We compared four supervised ML models: RF, SVM, XGBoost, and LightGBM. 
Proposed ML models, trained on a domain-specific dataset of IEEE 802.11 and LTE wireless networks, were evaluated not only on a test set from the same domain but also on unseen data from a different engineering domain - an Airport transportation system.
On the IEEE 802.11 and LTE test set, the ML models achieve F1-scores of up to 99.44\% and 98.34\% for the Plan and Absorb stages, respectively, while the more challenging Recover stage reaches 74.34\%.
On the independent Airport dataset, F1-scores range from 89.8\% to 94.7\% for Plan, 85.3\% to 93.4\% for Absorb, and 83.7\% to 91.3\% for Recover. Notably, Recover-stage classification improves by about 11–30 percentage points across individual models compared with the IEEE 802.11 and LTE test sets. 
The key overall resilience metrics, $R_t$ and $RL$, have APE below 3\% in 95.7\% and 100\% of cases, respectively, on the IEEE 802.11 and LTE test sets. On the Airport dataset, $RL$ maintains APE below 3\% in 87\% of cases, while $R_t$ achieves APE below 10\% in 64.8\% of cases.
We developed the proposed ML-based methods for resilience stage recognition and resilience quantification using a purpose-built dataset comprising system performance measurements from IEEE 802.11 WLAN and LTE communication systems operating under disruptive conditions. The developed methods were subsequently validated on a dataset from an airport transportation system, representing a different operational domain. We used the Airport dataset exclusively for external validation, without additional training or fine-tuning. To the best of our knowledge, this provides the first benchmark dataset capturing system-performance trajectories consistent with the resilience curve archetype for supervised stage recognition. The results indicate the proposed approach's potential for cross-domain transferability and generalization in system resilience analysis.

The study also has limitations that motivate future work. Both datasets are simulation-based and limited to single disruptions after which performance returns to its pre-event level, so the Adapt stage was incorporated into the Plan class rather than modeled separately. Although detecting the Recover stage is more challenging in wireless network data, the Recover-stage precision exceeds 99\% on the Airport dataset. Future work will validate the approach on real-world monitoring data and other infrastructure domains (e.g., power, water, and cyber-physical systems), consider multiple and compound disruptions, explicitly model the Adapt stage, and explore sequence models (e.g., recurrent or temporal convolutional networks) and adaptive windowing around stage transitions to improve Recover-stage recognition.

\section*{Acknowledgment}
The dataset generated and used in this study is publicly available in the Zenodo repository \cite{strzoda2026recognition}.

\bibliographystyle{IEEEtran}
\bibliography{bibliography}

\begin{IEEEbiography}
[{\includegraphics[width=1in,height=1.25in,clip,keepaspectratio]{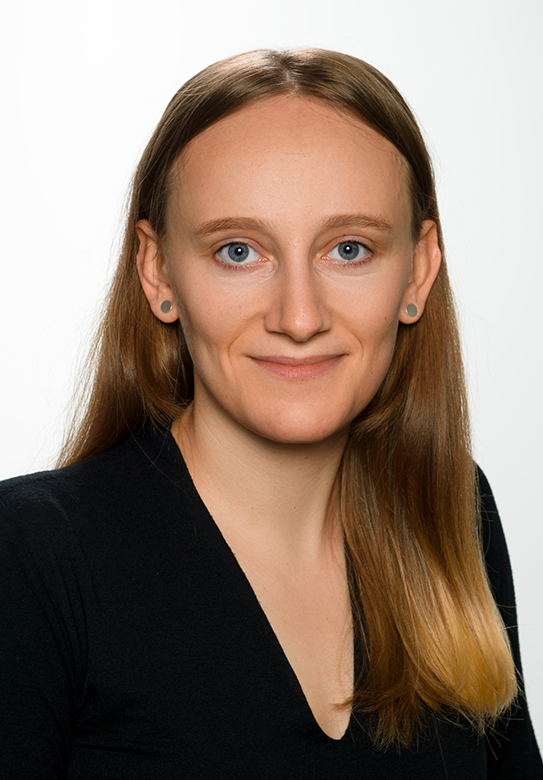}}]{Anna Strzoda}
is with the Institute of Theoretical and Applied Informatics, Polish Academy of Sciences. She investigates methods for resilience analysis of engineering systems within the NATO SPS Program (MYP G7979). Her research interests focus on the design, optimization, analysis, and resilience of wireless communication systems. Her research addresses a broad range of problems in wireless communication systems, including long-range (e.g., LPWAN) and short-range (e.g., UWB, BLE) networks, with particular emphasis on indoor and outdoor localization, radio network optimization, data analysis, and signal processing. Her work also covers IoT and Smart City applications and the use of ML models in wireless networks. 
\end{IEEEbiography}
\vspace*{-3pt}
\begin{IEEEbiography}
[{\includegraphics[width=1in,height=1.25in,clip,keepaspectratio]{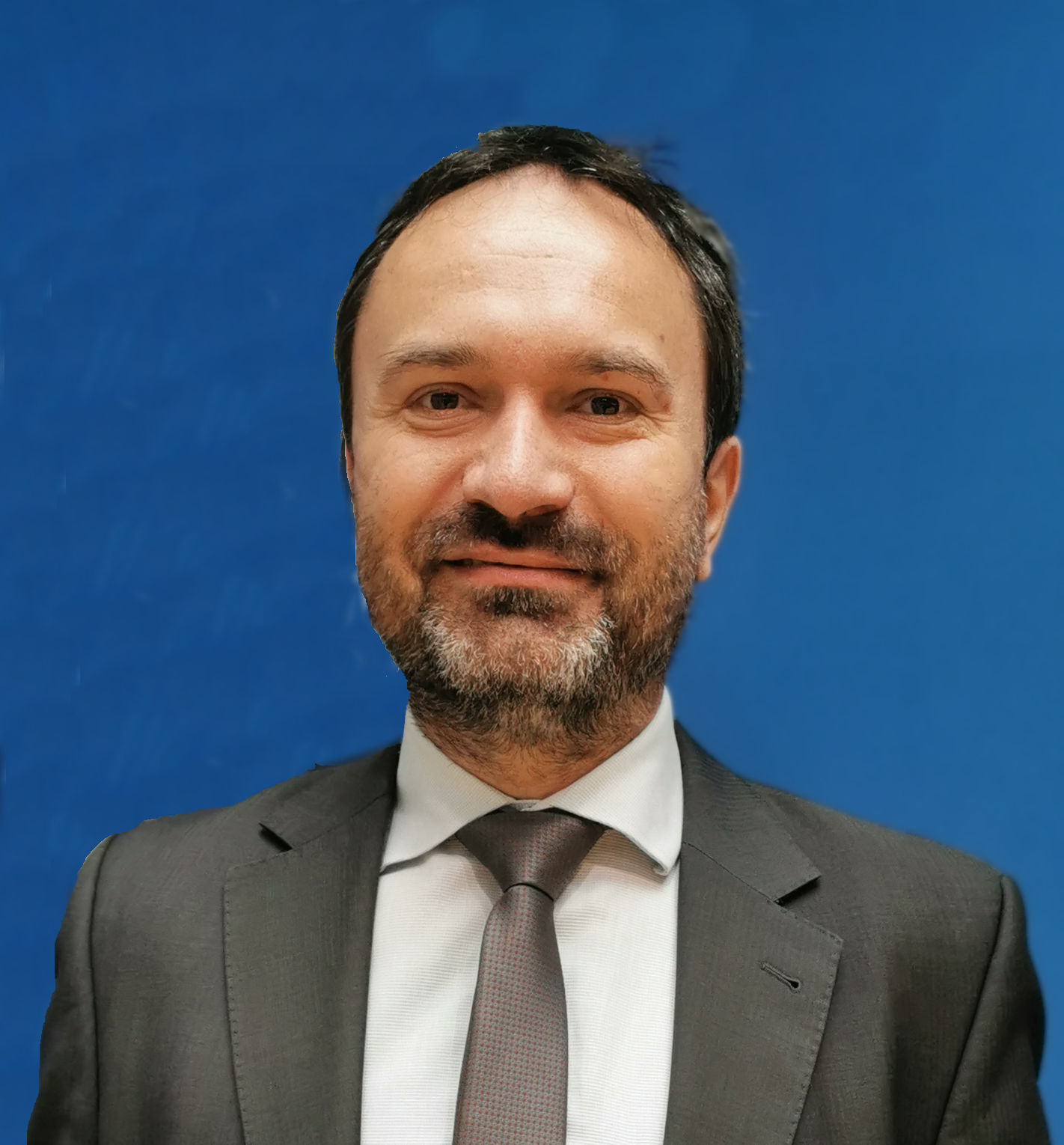}}]{Krzysztof Grochla}
is a professor at IITiS PAN; since 2023, he has served as the Director of the Institute. He acquired his habilitation in 2020 from Silesian University of Technology. He has authored over 100 research papers and is the inventor of three patents granted by USPTO. He works on resilience analysis of interconnected systems and critical infrastructure within the projects Infrastructure Recovery through Resilience Stress Testing in Ukraine (2023/05/Y/ST7/00192) and NATO SPS Program (MYP G7979). His research is oriented toward the performance evaluation of networking protocols and mechanisms for autoconfiguration and optimization of wireless networks. He participates in the standardization work of CEN/TC 294/WG 7 and the LoRa Alliance. 
\end{IEEEbiography}
\vspace*{-3pt}
\begin{IEEEbiography}
[{\includegraphics[width=1in,height=1.25in,clip,keepaspectratio]{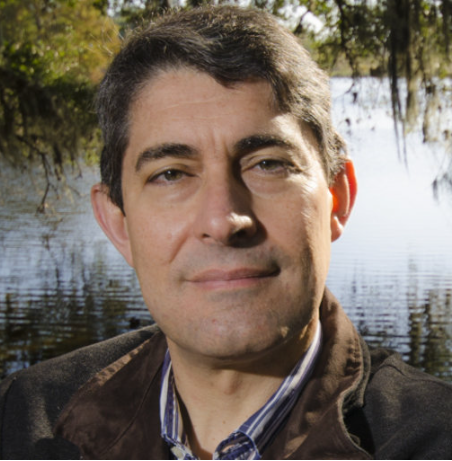}}]{Rafael Muñoz-Carpena}
is a Distinguished Professor specializing in environmental modeling and hydrology. His research applies systems approaches to water and environmental management, focusing on the resilience of engineered and natural systems. He develops mathematical models and methods for uncertainty analysis, model evaluation, and goodness-of-fit assessment. His work combines advanced computational tools with practical decision support for complex environmental systems. He is a Fellow of the American Society of Agricultural and Biological Engineers and the American Association for the Advancement of Science.
\end{IEEEbiography}

\end{document}